\documentclass{article}

\usepackage{iclr2027_conference,times}

\usepackage[utf8]{inputenc} 
\usepackage[T1]{fontenc}    
\usepackage[hidelinks]{hyperref} 
\usepackage{url}            
\usepackage{booktabs}       
\usepackage{amsfonts}       
\usepackage{nicefrac}       
\usepackage{microtype}      
\usepackage{enumitem}
\usepackage{booktabs,multirow,graphicx,amsthm,array}
\usepackage{mathrsfs}
\usepackage{amssymb}
\usepackage[table]{xcolor}
\usepackage{caption}
\usepackage{subcaption}
\usepackage[toc,page]{appendix}

\usepackage{amsmath,amsfonts,bm}

\def\figref#1{figure~\ref{#1}}

\def\1{\bm{1}}

\DeclareMathAlphabet{\mathsfit}{\encodingdefault}{\sfdefault}{m}{sl}
\SetMathAlphabet{\mathsfit}{bold}{\encodingdefault}{\sfdefault}{bx}{n}

\newcommand{\tworow}[2]{\begin{tabular}[c]{@{}c@{}}#1\vspace{-2pt}\\#2\end{tabular}}

\newcommand{\method}{Velox}
\newif\ifrevhl \revhlfalse
\newcommand{\rev}[1]{\ifrevhl{\color{blue}#1}\else#1\fi}
\newcommand{\revblue}{\ifrevhl\color{blue}\fi}
\definecolor{linkblue}{RGB}{0,82,163}

\title{Velox: Spatial Memory Makes\\World Models Efficient}

\author{%
  Qiuhong Shen \qquad Shizun Wang \qquad Xinchao Wang\thanks{Corresponding author.} \\[3pt]
  \normalfont National University of Singapore
}

\hypersetup{
  pdftitle={Velox: Spatial Memory Makes World Models Efficient},
  pdfauthor={Qiuhong Shen, Shizun Wang, Xinchao Wang}
}

\iclrfinalcopy

\ificlrfinal
\makeatletter
\def\@maketitle{\vbox{\hsize\textwidth\centering
  {\LARGE\sc \@title\par}
  \vskip 0.25in
  \begin{tabular}[t]{c}\bf\@author\end{tabular}%
  \vskip 0.3in minus 0.1in}}
\makeatother
\fi

\begin{document}

\maketitle
\ificlrfinal\lhead{Preprint}\fi

\begin{abstract}
Video world models turn actions and camera motion into long visual rollouts, making video a natural interface for interactive simulation.
Such interaction requires responsive generation, yet coherent rollouts must retain visual memory of what has already been generated. Chunked video diffusion models usually keep this context in growing KV caches, so later chunks become progressively slower. To address this bottleneck, we introduce \method{}, a training-free attention router for region-level context selection using frustum slab correspondences from streaming geometry. \method{} represents historical blocks as camera frustum slabs with depth intervals and obtains current-block depth through recency-priority multiwarp. For well-covered regions, frustum slab overlap softly modulates each layer's model-derived block importance. Uncovered regions retain the model-derived route without geometric bias.
The fixed token-block granularity matches FlashAttention-style tiled execution and reduces the attention term that grows with rollout length. Frustum slab construction runs asynchronously with denoising on the same H200. At chunk 48, \method{} accelerates attention by $4.36\times$ and denoising by $2.43\times$. On the full 158-case WBench navigation split, it remains within $0.15$ API-free average points of dense LingBot-Fast-Cam.
Please see our \href{https://velox-world.com/velox/}{\textcolor{linkblue}{project page}} for more details.
\end{abstract}

\section{Introduction}

\label{sec:introduction}

Video world models are emerging as a practical interface for interactive simulation. A user specifies an action trajectory or camera motion, and the model rolls the world forward by generating future video frames. This direction builds on rapid progress in video diffusion transformers and large video foundation models~\citep{ho2020denoising,rombach2022high,peebles2023scalable,blattmann2023stable,chen2024videocrafter2,yang2024cogvideox,kong2024hunyuanvideo,wan2025wan,ma2025step}. Recent works increasingly target controllable long video worlds rather than isolated short clips~\citep{alonso2024diffusion,valevski2024diffusion,bruce2024genie,parkerholder2024genie2,teng2025magi,gao2026advancing,cao2026hyworld}. For these models, context management is a central methodological challenge. Past frames carry the appearance, geometry, and object state that future frames must respect. Without access to the right history, long rollouts drift, forget revisited regions, or re-synthesize the same scene inconsistently.

The most direct way to preserve this history is to retain an ever-growing visual context. Chunked or causal video diffusion models let each new chunk attend to historical frame tokens through a KV cache. Dense FlashAttention makes this practical at moderate lengths~\citep{dao2022flashattention,dao2023flashattention}, and causal diffusion or self-forcing training makes long rollouts natural~\citep{chen2024diffusion,huang2025self}. The cost, however, grows with the number of generated chunks. Later chunks inherit larger context even when the current camera overlaps only a small part of the past scene. As a result, chunk-wise generation becomes progressively slower exactly when long-horizon consistency matters most.

\begin{figure*}[h!]
    \vspace{-2mm}
    \centering
    \includegraphics[width=\textwidth]{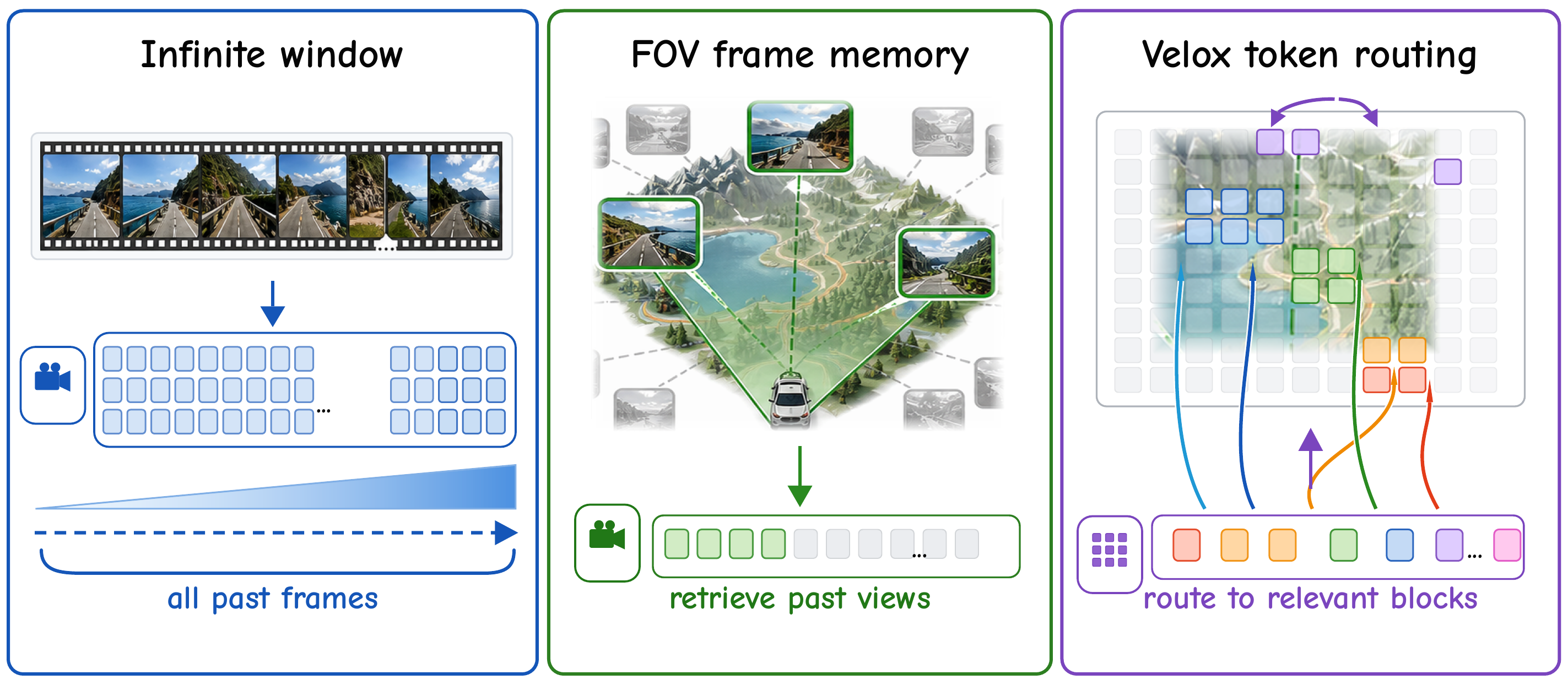}
    \caption{\textbf{Three context-management paradigms for long-horizon world generation.} Dense chunked generation keeps an ever-growing visual context. Frame-memory methods retrieve a small set of past views using camera overlap. \method{} routes each current latent region to compact, geometry-relevant history blocks using frustum slab correspondences.}
    \label{fig:context-paradigms}
\end{figure*}

This motivates context selection rather than naive context accumulation. Frame- and view-memory methods select a smaller set of relevant past observations. Context-as-Memory retrieves historical views by camera field-of-view overlap, and VMem stores surfel-indexed views for consistent revisits~\citep{yu2025context,li2025vmem}. AnchorWeave and related works~\citep{xiao2025worldmem,wu2025spatialmemory,wang2026anchorweave} further inject warped renderings from persistent memory through ControlNet~\citep{zhang2023adding} to stabilize camera-controlled generation. These methods show that not all past frames are equally useful, and \figref{fig:context-paradigms} summarizes this progression from dense history to view-level retrieval. Yet frame-level selection remains coarse for transformer attention. It provides the same selected context to all regions in the active chunk, although dynamic scenes often need local temporal continuity and different regions in the same noisy chunk can depend on different parts of the past. Methods that inject warped point renderings into the model input can also require architectural adaptation or fine-tuning, which limits direct reuse of a pretrained video generator.

\vspace{-4mm}

Inspired by inverse rendering works~\citep{kerbl20233d,mildenhall2021nerf}, we observe that regions whose geometry has already been observed should primarily attend to their corresponding past regions. Newly revealed or uncertain regions, in contrast, may require model-derived semantic context and local temporal neighbors. A road block, a facade block, and newly revealed sky in the same noisy chunk should therefore not necessarily attend to the same past views. The question, then, is no longer only which frames to retrieve, but whether long-context video generation can assign each region its own relevant history while preserving local motion continuity and leaving the pretrained generator unchanged.

We consider this question at the token-block level in the block-wise execution model of FlashAttention~\citep{dao2023flashattention}. Our key insight is that camera frustum slabs with depth intervals can provide region-specific routing signals at the same granularity used by tiled attention. Instead of adding a new conditioning branch or fine-tuning the video generator, \method{} summarizes generated history as per-block frustum slabs and uses their overlap to guide which historical KV token blocks each current region should see. For each target chunk, \method{} obtains query-side depth through recency-priority multiwarp and scores query and historical blocks by world-space OBB frustum slab overlap. In well-covered regions, this score softly reweights each layer's model-derived block importance before selection. Uncovered regions retain the model-derived route without geometric bias. Previous-chunk retention and global-aligned Gilbert neighborhoods preserve local motion continuity\rev{, including across chunk boundaries}.

Across 48-chunk timing traces and the 158-case WBench navigation split~\citep{ying2026wbench}, \method{} reduces the growing long-context attention cost while preserving aggregate quality near dense attention. Its advantage grows with the number of generated chunks. At chunk 48, attention latency decreases from approximately $34.2$ ms with dense FlashAttention to $7.8$ ms, a $4.36\times$ speedup. Denoising time decreases from approximately $7.19$ s to $2.96$ s under the same single-H200 asynchronous setting. Built on LingBot-Fast-Cam~\citep{gao2026advancing}, \method{} remains within $0.15$ points of dense attention on the WBench API-free average \rev{and stays level on the 19-metric overall average ($75.60$ versus $75.52$), a gap we treat as parity.} These results support the scoped claim that soft frustum slab routing can alleviate the late-chunk attention bottleneck \rev{while keeping benchmark quality close to dense attention}.

Our contributions can be summarized as:
\begin{itemize}[leftmargin=*,itemsep=2pt,topsep=2pt]
    \item We introduce \method{}, a training-free framework that uses frustum slab correspondences to reduce growing KV attention cost in chunked video world models.
    \item We devise a soft geometric fusion rule that uses OBB frustum slab overlap to modulate each layer's model-derived block importance while preserving an unbiased fallback for uncovered regions.
    \item We implement \method{} with global-aligned Gilbert KV storage, an optimized sparse attention kernel, and asynchronous frustum slab construction on the same H200 as denoising, reaching a $4.36\times$ attention speedup and a $2.43\times$ denoising speedup at chunk 48.
\end{itemize}

\section{Related Work}

\subsection{Video World Models and Spatial Memory}
Video generation has progressed from latent diffusion to scalable diffusion transformers and rectified flow~\citep{ho2020denoising,rombach2022high,blattmann2023stable,peebles2023scalable,esser2024scaling}, with foundation models supplying strong visual priors~\citep{yang2024cogvideox,kong2024hunyuanvideo,wan2025wan,hacohen2024ltx,ma2025step,gao2025seedance,team2025longcat}. Long-video methods extend these priors through progressive generation, self-rollout training, and causal chunking~\citep{kim2024fifo,jin2024pyramidal,henschel2024streamingt2v,xie2024progressive,gu2025long,chen2024diffusion,huang2025self,teng2025magi,chen2025skyreels,liu2025infinitystar}, while interactive models target controllable persistent worlds~\citep{bruce2024genie,parkerholder2024genie2,alonso2024diffusion,valevski2024diffusion,feng2024matrix,zhang2025matrixgame,che2024gamegen,yang2024playable,yu2025gamefactory,oasis2024,gao2026advancing,cao2026hyworld,shen2026world}. These directions make history management central: more context supports consistency but makes dense attention increasingly expensive. Spatial-memory methods address this tension by retrieving views, preserving revisits, or expanding reconstructed worlds~\citep{yu2025context,li2025vmem,xiao2025worldmem,wu2025spatialmemory,huang2025voyager,chen2025flexworld,yu2024wonderjourney,yu2024wonderworld,engstler2024invisible}. Camera-aware and controllable generation~\citep{yu2024viewcrafter,yu2025trajectorycrafter,bai2025recammaster,he2024cameractrl,he2025cameractrl,yang2025omnicam,ren2025gen3c,yu2026vifeedit} and streaming reconstruction~\citep{wang2024dust3r,mast3r_eccv24,cabon2025must3r,Yang_2025_Fast3R,wang2024vggsfm,wang2025vggt,wang2025continuous,keetha2025mapanything} further make projectable memory practical. \method{} uses this memory to route pretrained attention rather than adding a conditioning pathway or retrained memory module.

\subsection{Efficient Attention and Context Routing}
Long-context transformers reduce attention cost through sparsity, locality, compression, state reuse, heavy-hitter selection, and KV eviction~\citep{kitaev2020reformer,zaheer2020big,beltagy2020longformer,wang2020linformer,dai2019transformer,liu2023ring,lee2025infinitehip,jiang2024minference,li2024snapkv,zhang2023h2o,kwon2023efficient,li2026jeva}, while FlashAttention is the dense IO-aware baseline~\citep{dao2022flashattention,dao2023flashattention}. Video accelerators additionally exploit spatial, temporal, denoising-step, and feature regularity through adaptive sparsity, caching, and quantized attention~\citep{zhang2025fast,xi2025sparse,yang2025sparsevideogen2,zhang2025spargeattn,xia2025training,ding2025efficientvditefficientvideodiffusion,zhang2024sageattention,zhang2025accvideo,liu2024clear,liu2025region,liu2024timestep,guo2024blocksparse,nawaz2026worldcache,feng2026worldcache}. These approaches usually derive sparsity from position, cached features, or attention statistics, whereas frame-memory methods select views but retain dense attention within them. \method{} connects both directions by using projective correspondence as a soft prior for region-specific attention routes, with model-derived fallback and temporal safeguards.

\section{Method}
\label{sec:method}

\begin{figure*}[h!]
    \vspace{-2mm}
    \centering
    \includegraphics[width=\textwidth]{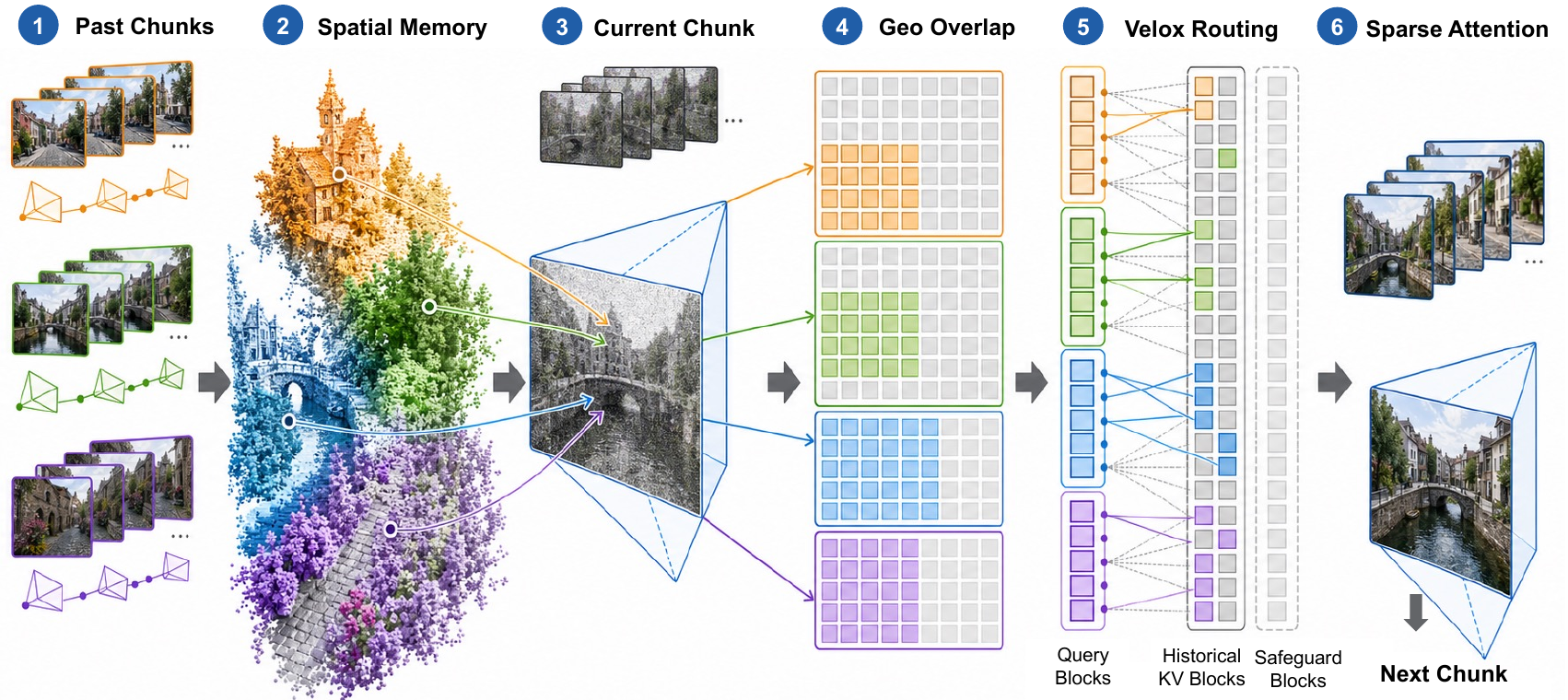}
    \caption{\textbf{Velox overview.} Velox leaves the pretrained chunked video generator unchanged and replaces dense historical attention with frustum slab routed attention. Historical chunks are summarized by per-block frustum slabs with depth intervals, target query blocks obtain depth through multiwarp, and block routes softly fuse frustum slab overlap with layer-specific model importance before adding temporal and global Gilbert safeguards. Indexed sparse attention then produces the next chunk.}
    \label{fig:velox-overview}
\vspace{-4mm}
\end{figure*}

\noindent\textbf{Overview.}
Long-horizon generation requires persistent visual history, yet dense KV attention makes every current region process all retained blocks. \method{} keeps the pretrained generator and conditioning interface unchanged and modifies only the inference-time visibility relation between current query blocks and historical KV blocks. It represents history as camera frustum slabs with depth intervals, uses their overlap as a soft prior over each layer's model-derived block importance, and exposes each current region to a compact route. Figure~\ref{fig:velox-overview} summarizes this training-free pipeline. \rev{In this section, we first define chunked generation and the token block that serves as the routing unit (Section~\ref{subsec:method-prelim}). We then describe how frustum slabs give each pair of current and historical blocks a geometric score and a coverage test (Section~\ref{subsec:method-correspondence}). Next, we show how this score softly guides the route of each layer (Section~\ref{subsec:method-routing}). Finally, we analyze the cost of the route (Section~\ref{subsec:method-complexity}).}

\subsection{Preliminaries: Chunked World Generation}
\label{subsec:method-prelim}

Consider a video world model that generates chunks $\bm{x}_0,\bm{x}_1,\ldots$ autoregressively. At denoising step $s$, chunk $c$ has noisy latent tokens $\bm{z}_c \in \mathbb{R}^{T \times H \times W \times d}$, text/action controls $(y_c,a_c)$, and a camera sequence $\pi_c=\{\pi_{c,t}\}_{t=1}^{T}$, where $T$ is the number of latent frames, $H\times W$ is the latent spatial grid, $d$ is the channel dimension, and each $\pi_{c,t}$ maps 3D scene locations to the corresponding latent/image view. A pretrained diffusion transformer with parameters $\theta$ predicts
\begin{equation}
    \hat{\bm{\epsilon}}_c =
    f_\theta\!\left(\bm{z}_c, s, y_c, a_c, \pi_c,
    \mathcal{C}_{<c}\right),
    \label{eq:chunk-model}
\end{equation}
where $\mathcal{C}_{<c}$ is the retained historical context, implemented as a visual KV cache in the transformer. If $\mathcal{C}_{<c}$ contains all previous chunks, the attention cost grows with history length even when only a small part of the past scene is visible from the current camera.

We operate at block granularity. Let $\sigma$ be the global-aligned Gilbert order, and let $\mathcal{B}_c=\{B_i\}_{i=1}^{N}$ partition the current chunk into blocks of $b_{\mathrm{tok}}$ tokens. Historical KV blocks use the same size, and $M_c$ denotes their visible count before chunk $c$. For fixed chunk $c$ and layer $\ell$, write $Q_i$ for current block $B_i$ and $(K_r,V_r)$ for visible KV block $r\in\{1,\ldots,M_c\}$.

This block size matches FlashAttention-style execution~\citep{dao2022flashattention,dao2023flashattention}, which streams fixed query and KV tiles. Abstractly, history attention decomposes into
\begin{equation}
    \sum_{i=1}^{N}\sum_{r=1}^{M_c}
    \operatorname{TileAttn}\!\left(Q_i,K_r,V_r\right),
    \label{eq:flash-tile}
\end{equation}
where each interaction computes $Q_iK_r^\top$, updates online softmax, and applies $V_r$. Dense FlashAttention still schedules every current-history pair. A fixed route removes entire KV tiles before launch while preserving block-contiguous gathers. Token-level or variable-size sparsity would require irregular row-wise indexing.

The active attention window is ordered by global-aligned Gilbert rank. Dense visual attention for block $i$ uses every visible block:
\begin{equation}
    \mathrm{Attn}_{\mathrm{dense}}\!\left(Q_i\right)
    =
    \mathrm{softmax}\!\left(
    Q_i K_{1:M_c}^\top / \sqrt{d}
    \right)V_{1:M_c}.
    \label{eq:dense-attn}
\end{equation}
\method{} replaces $1{:}M_c$ by a route $\mathcal{R}_{i}^{\ell}\subseteq\{1,\ldots,M_c\}$ at each transformer layer $\ell$. The layer index matters because every layer produces its own pooled query-key importance distribution, while the geometric correspondence is shared across layers within a denoising step.

\subsection{Frustum Slab Correspondence}
\label{subsec:method-correspondence}

\method{} does not route by brittle reconstructed sample identity. Instead, it stores each historical KV block as one or more \emph{frustum slab primitives}, each comprising a source-camera token rectangle and robust depth interval:
\begin{equation}
    \psi_{j,r,m} =
    \left(
    r,\ t,\ \mathbf{R}_{j,r,m},\ [z^{-}_{j,r,m},z^{+}_{j,r,m}],\
    \nu_{j,r,m}
    \right),
    \label{eq:frustum-state}
\end{equation}
Here $j$, $r$, $m$, and $t$ index the source chunk, global Gilbert block, primitive, and latent frame. The source-camera token rectangle is $\mathbf{R}_{j,r,m}$, its percentile depth interval is $[z^{-}_{j,r,m},z^{+}_{j,r,m}]$, and $\nu_{j,r,m}$ stores validity statistics. Blocks crossing frame or chunk boundaries have multiple primitives, aggregated by source chunk before scoring.

\noindent\textbf{Historical frustum slab construction.}
After chunk $j$ is decoded, the reconstruction backend produces depth, confidence, masks, and camera poses for the generated frames. \method{} transiently unprojects reliable depth samples into 3D, projects them into the chunk's latent cameras, buckets them by global Gilbert block, and computes a depth quantile ladder for each block fragment. Let $\operatorname{Quantile}_{p}(\cdot)$ denote the empirical $p$-th percentile over valid source-camera depths, and let $p_{\mathrm{lo}}$ and $p_{\mathrm{hi}}$ be fixed low and high percentile levels. The frustum slab interval is:
\begin{equation}
    z^{-}_{j,r,m}=\operatorname{Quantile}_{p_{\mathrm{lo}}}\!\left(\mathcal{Z}_{j,r,m}\right),
    \qquad
    z^{+}_{j,r,m}=\operatorname{Quantile}_{p_{\mathrm{hi}}}\!\left(\mathcal{Z}_{j,r,m}\right),
    \label{eq:frustum-quantile}
\end{equation}
where $\mathcal{Z}_{j,r,m}$ is the set of valid source-camera depths collected for that primitive. Validity gates require enough samples, enough covered tokens, bounded relative depth spread, and a minimum frustum slab thickness. This yields a compact frustum slab table for stored history without carrying dense reconstructed geometry in the routing score.

\noindent\textbf{Target-side multiwarp depth.}
To score a current query block, \method{} also needs a depth interval in the target camera. \rev{\method{} obtains} this Q-depth with recency-priority multiwarp. It caches the first and last pixel-frame depths of recent reconstructed chunks, then forward-warps all cached frames into the target chunk's latent cameras. For each target token $u$, let $f$ index cached source frames, let $n_f(u)$ be the number of warped depth samples that land on $u$, and let $k_{\mathrm{warp}}$ be the sample floor. The newest cached frame with at least $k_{\mathrm{warp}}$ samples wins:
\begin{equation}
    f^{\star}(u)
    =
    \max\{f \mid n_f(u) \ge k_{\mathrm{warp}}\},
    \qquad
    z_Q(u)=\operatorname{Quantile}_{50}\!\left(\mathcal{Z}_{f^{\star}(u),u}\right).
    \label{eq:multiwarp}
\end{equation}
Here $\mathcal{Z}_{f,u}$ is the depth sample set warped from cached frame $f$ to target token $u$, and $\operatorname{Quantile}_{50}$ is the median depth. If no cached frame reaches the sample floor, $f^{\star}(u)$ is undefined, the token has no Q-depth, and the row can fall back to model-derived routing. Intuitively, this rule favors the most recent reliable geometry, while still allowing revisits to be filled by older observations when the current camera returns to a previously seen surface. Appendix~\ref{app:multiwarp-qdepth} formalizes source filtering, target-camera reprojection, winner selection, and query-block aggregation.

\noindent\textbf{World-space frustum slab overlap.}
Given a query block $i$ with multiwarp depth and the historical primitives attached to KV block $r$, \method{} approximates each camera frustum slab by a camera-facing oriented bounding box. Let $\Psi_r=\{\psi_{j,r,m}\}$ be the primitive set attached to historical block $r$, and let $\mathcal{O}_i^Q$ and $\mathcal{O}_{j,r,m}^{K}$ denote the query and historical OBBs in world coordinates. The geometric score is:
\begin{equation}
    g_{i,r}
    =
    \max_{\psi_{j,r,m}\in\Psi_r}
    \frac{
    \operatorname{Vol}\!\left(
    \mathcal{O}_i^Q \cap \mathcal{O}_{j,r,m}^{K}
    \right)}
    {\operatorname{Vol}\!\left(\mathcal{O}_i^Q\right)
    +\operatorname{Vol}\!\left(\mathcal{O}_{j,r,m}^{K}\right)
    -\operatorname{Vol}\!\left(
    \mathcal{O}_i^Q \cap \mathcal{O}_{j,r,m}^{K}
    \right)+\epsilon}.
    \label{eq:frustum-iou}
\end{equation}
Here $\epsilon>0$ prevents division by zero. The implementation estimates the intersection with a deterministic OBB grid after a separating-axis test, then applies grouped fragment aggregation for multi-fragment KV columns. This score is view-aware because both the target and historical blocks carry camera orientation, token footprint, and depth extent. It also avoids tying correspondence to exact reconstructed sample identities, which are brittle under generated-video artifacts.

\noindent\textbf{Coverage.}
The router separates geometric score from geometric trust. A token is covered when its multiwarp depth lies inside at least one valid historical frustum slab that contains the token ray within a spatial and depth margin. The block coverage is:
\begin{equation}
    \rho_i =
    \frac{1}{|B_i|}
    \sum_{u\in B_i}
    \mathbf{1}\!\left[
    \exists r,\ \exists \psi_{j,r,m}\in \Psi_r:\ u \in \Pi_c(\psi_{j,r,m})
    \wedge z_Q(u)\in [z^{-}_{j,r,m},z^{+}_{j,r,m}]
    \right],
    \label{eq:coverage}
\end{equation}
where $\Pi_c(\psi_{j,r,m})$ denotes the projection footprint of the frustum slab primitive in the current target view, and $\eta_{\mathrm{cov}}$ is the coverage threshold. A block is covered when $\rho_i \ge \eta_{\mathrm{cov}}$. Thus $g_{i,r}$ identifies which historical KV blocks overlap the current frustum, while $\rho_i$ decides whether geometry is reliable enough to drive the route.

\subsection{Velox Attention from Frustum Slab Correspondence}
\label{subsec:method-routing}

The geometric score should guide a route without imposing the same hard block set on every layer. Reconstruction can leave Q-depth holes, stale warped depths can survive disocclusions, and different transformer layers can favor different visual evidence. A hard geometric top-$K$ route would override these layer-specific preferences. \method{} instead uses frustum slab overlap to modulate the model's own block importance.

\noindent\textbf{Layer-specific model importance.}
Let $\bar{q}_{i}^{\ell}$ be the mean query vector in block $i$ at layer $\ell$, and let $\bar{k}_{r}^{\ell}$ be the mean key vector in visible KV block $r$. We compute:
\begin{equation}
    p_{i,r}^{\ell}
    =
    \mathrm{softmax}_{r}
    \left(
    \bar{q}_{i}^{\ell}\left(\bar{k}_{r}^{\ell}\right)^{\top}/\sqrt{d}
    \right),
    \label{eq:model-prior}
\end{equation}
where the softmax is normalized over all visible KV blocks. This distribution preserves the pretrained transformer's layer-specific notion of useful context.

\noindent\textbf{Soft geometric fusion.}
For a well-covered query block with $\rho_i\geq\eta_{\mathrm{cov}}$, we first suppress overlap values at or below the noise threshold $\tau$ and rescale the remaining values:
\begin{equation}
    \widehat{g}_{i,r}
    =
    \mathbf{1}\!\left[g_{i,r}>\tau\right]
    \min\!\left(\frac{g_{i,r}}{g_{\mathrm{ref}}},1\right),
    \label{eq:soft-geo-score}
\end{equation}
where $g_{\mathrm{ref}}$ maps the observed frustum slab IoU range to a bounded boost. We then multiply the model importance by this geometric prior and renormalize:
\begin{equation}
    \widetilde{p}_{i,r}^{\ell}
    =
    \frac{
    p_{i,r}^{\ell}\left(1+\lambda\widehat{g}_{i,r}\right)}
    {\sum_{r'=1}^{M_c}
    p_{i,r'}^{\ell}\left(1+\lambda\widehat{g}_{i,r'}\right)}.
    \label{eq:soft-geo-fusion}
\end{equation}
Here $\lambda>0$ controls the fusion strength. A block with $g_{i,r}\geq g_{\mathrm{ref}}$ receives the maximum multiplier $1+\lambda$, while a block without vetted geometric overlap retains its original model importance. Because the multiplication acts on $p_{i,r}^{\ell}$, geometry can promote a model-supported correspondence without assigning every layer an identical route.

\noindent\textbf{Covered and uncovered routes.}
Let $\mathcal{P}_{c-1}$ denote the previous chunk's KV blocks. The route materializer reserves these blocks before selecting older history, so local temporal evidence consumes the same bounded budget as the long-range route. For well-covered blocks, \method{} fills the remaining $K_{\mathrm{cov}}-|\mathcal{P}_{c-1}|$ positions with the largest fused scores:
\begin{equation}
    \mathcal{R}_{i}^{\ell}
    =
    \mathcal{P}_{c-1}
    \cup
    \operatorname{Top}_{K_{\mathrm{cov}}-|\mathcal{P}_{c-1}|}
    \left(
    \{\widetilde{p}_{i,r}^{\ell}:r\notin\mathcal{P}_{c-1}\}
    \right),
    \qquad \rho_i\geq\eta_{\mathrm{cov}}.
    \label{eq:route-covered}
\end{equation}
For blocks with insufficient geometric coverage, \method{} leaves the model distribution unchanged. It sorts $\{p_{i,r}^{\ell}\}$ and takes the smallest prefix whose cumulative mass reaches $\eta_{\mathrm{imp}}$, subject to a route cap $K_{\mathrm{unc}}$:
\begin{equation}
    \mathcal{R}_{i}^{\ell}
    =
    \mathcal{P}_{c-1}
    \cup
    \operatorname{Prefix}_{\eta_{\mathrm{imp}},K_{\mathrm{unc}}-|\mathcal{P}_{c-1}|}
    \left(
    \{p_{i,r}^{\ell}:r\notin\mathcal{P}_{c-1}\}
    \right),
    \qquad \rho_i<\eta_{\mathrm{cov}}.
    \label{eq:route-uncovered}
\end{equation}
This branch lets the pretrained attention signal handle sky, newly exposed regions, and holes in the frustum slab table. The calibrated configuration uses $\tau=0.1$, $g_{\mathrm{ref}}=0.2$, $\lambda=2$, $\eta_{\mathrm{cov}}=0.9$, $K_{\mathrm{cov}}=80$, and $K_{\mathrm{unc}}=150$.

\noindent\textbf{Global-aligned Gilbert neighborhoods.}
\method{} preserves short-range video continuity through a global/window 26-neighbor adjacency over Gilbert blocks. \rev{This adjacency must stay valid across chunk boundaries, so \method{} builds a single global-aligned Gilbert block domain incrementally. It constructs the single-chunk Gilbert curve once and flips chunk-local coordinates according to chunk parity and the base-frame exit quadrant. As a result, every generated chunk matches the prefix of the full-sequence Gilbert curve, and neighbors across a chunk boundary become ordinary temporal neighbors in the visible window. Appendix~\ref{app:gilbert-order} gives the construction.}

Let $\mathcal{L}_{i}^{\mathrm{glob}}$ denote the visible 26-neighborhood of query block $i$ in this global Gilbert block domain, and let $\mathcal{F}_{0}$ denote optional reference-frame blocks. The materialized route is:
\begin{equation}
    \overline{\mathcal{R}}_{i}^{\ell}
    =
    \mathcal{R}_{i}^{\ell}
    \cup \mathcal{L}_{i}^{\mathrm{glob}}
    \cup \mathcal{F}_{0}.
    \label{eq:safeguard-route}
\end{equation}
The spatial neighbors and optional reference blocks are added as deterministic safeguards outside the route budget.
The sparse visual-attention output is
\begin{equation}
    \mathrm{Attn}_{\method}\!\left(Q_i\right)
    =
    \mathrm{softmax}\!\left(
    Q_i K_{\overline{\mathcal{R}}_{i}^{\ell}}^\top
    / \sqrt{d}
    \right)
    V_{\overline{\mathcal{R}}_{i}^{\ell}},
    \label{eq:velox-attn}
\end{equation}
at every transformer layer. With this formulation, \method{} preserves each layer's model-derived ranking, injects geometry only where correspondence is reliable, and still exposes the sparse kernel to a bounded block list.

\noindent\textbf{Gilbert KV layout.}
\method{} stores current and historical visual KV blocks in the same global-aligned Gilbert order used by the route. This keeps neighboring image regions close in memory and reduces per-call permutation overhead. The layout is an implementation optimization. The soft route in Eqs.~\eqref{eq:soft-geo-fusion} and~\eqref{eq:velox-attn} is independent of the particular space-filling curve, but global-aligned Gilbert makes local neighborhood preservation and block-contiguous KV storage share one coordinate frame.

\subsection{Complexity and Design Rationale}
\label{subsec:method-complexity}

The benefit of \method{} comes from separating two quantities that dense attention couples together: the amount of visual memory retained by the world model and the number of KV blocks exposed to each query block. With the fixed block size $b_{\mathrm{tok}}$, dense attention between $N$ current blocks and $M_c$ visible KV blocks has cost
\begin{equation}
    \mathcal{O}\!\left(N M_c b_{\mathrm{tok}}^2 d\right).
    \label{eq:dense-complexity}
\end{equation}
As more chunks are retained, $M_c$ grows even when the current camera needs only a small part of the past. \method{} keeps the memory available but caps the exposed route. Let $K_{\max}=\max(K_{\mathrm{cov}},K_{\mathrm{unc}})$ and let $\Delta_{\mathrm{safe}}=|\mathcal{L}_{i}^{\mathrm{glob}}\cup\mathcal{F}_{0}|$ denote the small add-on safeguard budget:
\begin{equation}
    \mathcal{O}\!\left(
    \sum_{i=1}^{N}
    |\overline{\mathcal{R}}_{i}^{\ell}|
    b_{\mathrm{tok}}^2 d
    \right)
    \approx
    \mathcal{O}\!\left(N(K_{\max}+\Delta_{\mathrm{safe}})b_{\mathrm{tok}}^2 d\right)
    \quad \text{per sparse attention layer.}
    \label{eq:velox-complexity}
\end{equation}
The approximation highlights the intended regime: the route and safeguard budgets are fixed, while $M_c$ grows with history length.

The same block unit also matches FlashAttention-style execution. Dense FlashAttention evaluates each query-key tile efficiently, but it still schedules all $N M_c$ tile pairs. \method{} changes the tile schedule to $\sum_i |\overline{\mathcal{R}}_{i}^{\ell}|$. A skipped block removes both the $QK^\top$ tile product and the KV tile load. Thus, routing reduces KV traffic when attention is memory-bound, and it reduces tile matrix products when attention is compute-bound. The calibrated route applies at every transformer layer, so the growing historical term is bounded throughout the denoising network.

\section{Experiments}
\label{sec:experiments}

\subsection{Implementation Details}
\label{subsec:exp-impl}

We instantiate \method{} on the LingBot-Fast-Cam inference stack~\citep{gao2026advancing} with a Wan-style video diffusion transformer. The reported configuration uses recency-priority multiwarp Q-depth, world-space OBB frustum slab IoU, grouped fragment aggregation, and global-aligned Gilbert KV storage. We apply fixed soft geometric fusion to well-covered rows at every transformer layer, with $\tau=0.1$, $g_{\mathrm{ref}}=0.2$, $\lambda=2$, and $\eta_{\mathrm{cov}}=0.9$. Covered rows retain $K_{\mathrm{cov}}=80$ blocks. Uncovered rows use the model-derived cumulative-mass route with $K_{\mathrm{unc}}=150$ and no route-size floor.

The measured sparse path is a native Triton indexed block-sparse attention implementation rather than a dense call with a Python mask. Current queries are reordered once per chunk, KV caches are stored in global-aligned Gilbert order, and the kernel consumes routed block lists with a runtime route-length bound. The previous chunk is always retained within the route budget, while current-chunk local neighbors and reference-frame blocks act as deterministic safeguards. Frustum slab state is prepared asynchronously with a two-chunk freshness lag, so chunk $c$ normally consumes geometry through chunk $c-2$. In the reported timing traces, DiT denoising and asynchronous geometry construction run concurrently on one H200.

We evaluate quality on all 158 cases of the WBench navigation split~\citep{ying2026wbench}. The three LingBot-based rows use identical generated cases and the Gemini judge for VLM-scored metrics. Published WBench rows use the official Doubao judge, so judge-dependent values should not be compared across these two groups. We therefore include an API-free average that excludes VLM-scored setting metrics and is comparable across judge backends.

\subsection{Attention Scaling over 48 Chunks}
\label{subsec:exp-speed}

\begin{figure}[ht!]
    \vspace{-3mm}
    \centering
    \begin{subfigure}[c]{0.48\linewidth}
        \centering
        \includegraphics[width=\linewidth]{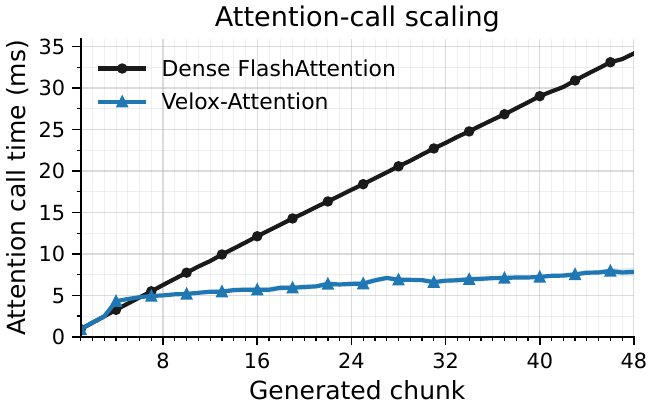}
        \caption{Per-call attention time.}
    \end{subfigure}
    \hfill
    \begin{subfigure}[c]{0.48\linewidth}
        \centering
        \includegraphics[width=\linewidth]{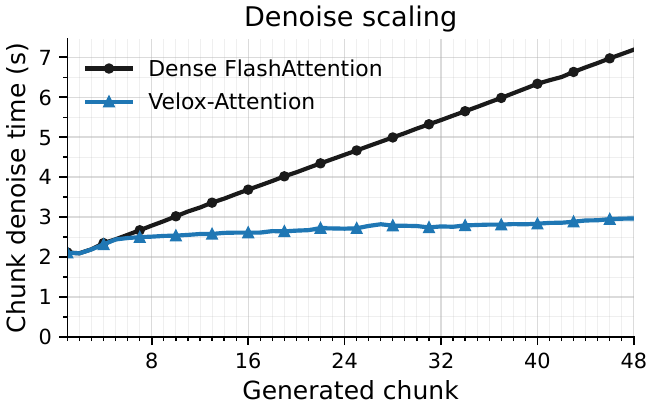}
        \caption{Chunk denoising time.}
    \end{subfigure}
    \caption{\textbf{Runtime scaling over a 48-chunk rollout.} The left plot measures per-call visual-attention latency. The right plot measures per-chunk DiT denoising time while asynchronous geometry construction shares the same H200. \method{} increasingly separates from dense FlashAttention as the retained context grows.}
    \label{fig:runtime-scaling}
\end{figure}

Figure~\ref{fig:runtime-scaling} reports 48-chunk timing traces. Dense FlashAttention grows with the retained context, while \method{} remains nearly flat because each query block receives a bounded route. At chunk 48, per-call attention latency decreases from approximately $34.2$ ms to $7.8$ ms, corresponding to a $4.36\times$ speedup. Under the same single-H200 asynchronous setting, chunk denoising time decreases from approximately $7.19$ s to $2.96$ s, corresponding to a $2.43\times$ speedup. The widening separation across the rollout shows that the gain stems from bounding the growing historical attention term rather than from a fixed short-context overhead. \rev{Note that these traces cover the attention call and the per-chunk denoising, and exclude model loading, VAE decoding, and the final drain of the geometry worker. Moreover, \method{} bounds the attention computed for each chunk but retains the full KV cache, so the cache memory still grows with the rollout as in dense attention. Appendix~\ref{app:runtime-path} details the scope of each measurement.}

\subsection{Comparison on WBench}
\label{subsec:exp-wbench}

Table~\ref{tab:wbench-navi} compares dense LingBot-Fast-Cam, model-only block-sparse routing, and \method{} under the same 158-case protocol. \method{} remains within $0.15$ points of dense attention on the API-free average \rev{and stays level on the 19-metric overall average ($75.60$ versus $75.52$). Given how small both gaps are, we regard them as parity with dense attention.} Against the block-sparse baseline, soft geometric fusion raises the API-free average from $73.11$ to $73.51$, the overall average from $75.01$ to $75.60$, and spatial consistency from $71.8$ to $74.3$. \rev{Notably, the averages move by less than one point, whereas spatial consistency rises by $2.5$ points.} \rev{These unpaired gaps suggest that geometry adds value beyond the matched sparse routing budget.} \rev{A paired test over the 158 cases (Appendix~\ref{app:paired}) finds no clear difference from dense attention on most metrics, with perspective consistency as the one material loss ($-4.7$ points, 95\% interval $[-8.6,-0.9]$). Quality thus stays near dense attention in aggregate.}

\begin{table}[ht!]
    \centering
    \scriptsize
    \caption{\textbf{WBench navigation results on 158 cases.} Scores are in $[0,100]$. Published baselines use Doubao, and the final three rows use Gemini for VLM-scored metrics. API-free excludes Setting, while Overall averages all 19 metrics.}
    \label{tab:wbench-navi}
    \setlength{\tabcolsep}{2.2pt}
    \resizebox{\textwidth}{!}{
    \begin{tabular}{l|cc|cccc|cccccccc}
        \toprule
        \multirow{2}{*}{Method}
        & \multicolumn{2}{c|}{Average}
        & \multirow[c]{2}{*}[-6pt]{\tworow{Quality}{Score}}
        & \multirow[c]{2}{*}[-6pt]{\tworow{Setting}{Adh.}}
        & \multirow[c]{2}{*}[-6pt]{\tworow{Interaction}{Adh.}}
        & \multirow[c]{2}{*}[-6pt]{\tworow{Physical}{Score}}
        & \multicolumn{8}{c}{Consistency} \\
        \cmidrule(lr){2-3}\cmidrule(lr){8-15}
        & \tworow{API-free}{Avg.}
        & \tworow{Overall}{Avg.}
        & & & &
        & \tworow{Back.}{Const.}
        & \tworow{Spatial}{Const.}
        & \tworow{Gated}{Spatial}
        & \tworow{Persp.}{Const.}
        & \tworow{Segment}{Cont.}
        & \tworow{Geom.}{Const.}
        & \tworow{Photo.}{Const.}
        & \tworow{Subject}{Const.} \\
        \midrule
        \multicolumn{15}{l}{\textit{Text-driven video generation models}} \\
        Seedance 1.5       & 73.03 & 79.68 & 82.1 & 82.9 & 68.0 & 68.4 & 89.6 & 72.7 & 72.4 & 70.5 & 96.2 & 82.4 & 76.8 & 90.1 \\
        Wan 2.7            & 72.35 & 80.76 & 81.5 & 91.4 & 66.0 & 71.8 & 89.4 & 71.0 & 71.0 & 78.2 & 92.4 & 83.7 & 76.4 & 90.7 \\
        Kling 3.0          & 74.03 & 81.53 & 81.4 & 91.0 & 70.3 & 69.3 & 92.3 & 75.2 & 75.1 & 76.8 & 93.0 & 88.9 & 79.9 & 88.5 \\
        YUME 1.5           & 71.87 & 76.51 & 77.7 & 72.4 & 72.0 & 65.2 & 90.3 & 71.5 & 71.4 & 48.0 & 99.4 & 88.0 & 83.3 & 88.8 \\
        HY-Video 1.5       & 74.12 & 81.16 & 77.6 & 85.6 & 71.8 & 67.4 & 92.1 & 79.2 & 75.1 & 86.6 & 99.4 & 94.6 & 80.3 & 91.6 \\
        LTX 2.3            & 69.42 & 76.21 & 77.1 & 85.2 & 67.6 & 64.9 & 88.3 & 70.2 & 70.2 & 69.8 & 75.8 & 76.9 & 79.2 & 87.2 \\
        LongCat-Video      & 71.83 & 78.65 & 75.4 & 72.3 & 63.1 & 68.9 & 95.1 & 83.3 & 66.2 & 81.5 & 99.4 & 95.4 & 82.2 & 93.4 \\
        Kairos 3.0         & 69.93 & 75.34 & 74.0 & 70.3 & 65.1 & 60.4 & 91.1 & 76.8 & 62.0 & 76.3 & 94.3 & 89.0 & 80.8 & 90.8 \\
        \midrule
        \multicolumn{15}{l}{\textit{Camera-controlled world models}} \\
        SANA-WM            & 74.65 & 77.86 & 79.3 & 76.1 & 82.1 & 61.9 & 90.5 & 77.1 & 76.4 & 49.0 & 97.5 & 88.7 & 81.2 & 85.4 \\
        HY-World 1.5       & 77.78 & 80.43 & 78.1 & 72.2 & 87.5 & 66.3 & 92.7 & 90.6 & 84.9 & 62.5 & 100.0 & 92.0 & 83.1 & 89.1 \\
        Fantasy-World      & 72.65 & 77.58 & 72.4 & 71.3 & 72.1 & 66.8 & 94.2 & 80.6 & 64.2 & 79.8 & 100.0 & 95.3 & 84.8 & 92.5 \\
        InSpatio-World     & 73.95 & 77.99 & 71.5 & 71.4 & 72.8 & 65.2 & 95.0 & 93.8 & 66.5 & 72.5 & 100.0 & 97.3 & 87.4 & 94.4 \\
        Astra              & 65.67 & 67.31 & 67.1 & 59.6 & 67.7 & 51.4 & 85.3 & 64.7 & 63.3 & 30.0 & 86.6 & 85.6 & 87.5 & 83.5 \\
        \midrule
        \multicolumn{15}{l}{\textit{Ours built on LingBot-Fast-Cam (Gemini judge)}} \\
        LingBot-Fast-Cam~\citep{gao2026advancing} & \textbf{73.66} & 75.52 & 77.7 & 79.9 & \textbf{79.4} & \textbf{39.0} & 87.9 & 73.0 & 73.0 & \textbf{81.7} & 90.5 & \textbf{80.5} & 77.6 & \textbf{87.2} \\
        Block-Sparse       & 73.11 & 75.01 & 77.9 & 79.1 & 78.0 & 38.6 & 87.6 & 71.8 & 71.8 & 75.7 & \textbf{94.3} & 80.4 & 77.1 & 85.8 \\
        \textbf{\method{}} & 73.51 & \textbf{75.60} & \textbf{78.1} & \textbf{80.3} & 78.8 & 38.8 & \textbf{88.0} & \textbf{74.3} & \textbf{74.3} & 77.0 & 92.4 & \textbf{80.5} & \textbf{78.0} & 86.4 \\
        \bottomrule
    \end{tabular}
    }
    \vspace{-2mm}
\end{table}

\subsection{Geometry and Model-Importance Agreement}
\label{subsec:exp-recall}

Soft fusion assumes that frustum slab overlap identifies useful history without imposing a hard replacement set. We audit this relation using offline affinity dumps. For layer $\ell$, threshold $\tau$, and well-covered query block $i$, let $\mathcal{H}_i$ be the $K_{\mathrm{hist}}$ candidate history blocks after removing the self block, $\mathcal{G}_i(\tau)=\{h\in\mathcal{H}_i:g_{i,h}>\tau\}$ the $n_i(\tau)$ positive-overlap blocks, and $\mathcal{P}_i^{50}(\ell)$ the model-preferred top-50 after averaging affinity over batch and heads.

With $a_i(\ell,\tau)=|\mathcal{G}_i(\tau)\cap\mathcal{P}_i^{50}(\ell)|$, geometric recall is $R_{\mathrm{geo}}(\ell,\tau)=\mathbb{E}_i[a_i(\ell,\tau)/n_i(\tau)]$, against the random expectation $50/143=0.350$. The fixed-budget set $\mathcal{U}_i^{50}(\ell,\tau)$ contains all positive-overlap blocks and the highest-importance outside blocks needed to reach 50. Its agreement is $R_{50}(\ell,\tau)=\mathbb{E}_i[|\mathcal{P}_i^{50}(\ell)\cap\mathcal{U}_i^{50}(\ell,\tau)|/50]$. We omit rows with no positive overlap or non-finite scores and average over rows with at least $0.9$ coverage from the last chunk of four calibration scenes and four denoising steps.

At $\tau=0.1$, Table~\ref{tab:recall-audit} shows agreement above the random expectation, with layer 14 reaching $R_{\mathrm{geo}}=0.7302$ and $R_{50}=0.8909$. The layer variation motivates retaining each model-derived distribution instead of imposing a layer-invariant geometric set. \rev{Appendix~\ref{app:soft-calibration} uses the same statistics to set $\lambda$ and $g_{\mathrm{ref}}$, so fusion adds a measured geometric bias without replacing the model ranking.}

\begin{table}[ht!]
    \centering
    \scriptsize
    \caption{\textbf{Geometry and model-importance agreement at $\tau=0.1$.} $R_{\mathrm{geo}}$, lift over the random expectation, and fixed-budget overlap $R_{50}$ are defined in the text.}
    \label{tab:recall-audit}
    \setlength{\tabcolsep}{5pt}
    \begin{tabular}{lccc}
        \toprule
        Layer & $R_{\mathrm{geo}}$ & Lift & $R_{50}$ \\
        \midrule
        10 & 0.6812 & +0.3315 & 0.8723 \\
        14 & \textbf{0.7302} & \textbf{+0.3806} & \textbf{0.8909} \\
        19 & 0.6929 & +0.3433 & 0.8770 \\
        23 & 0.6729 & +0.3233 & 0.8716 \\
        24 & 0.6848 & +0.3351 & 0.8739 \\
        25 & 0.6652 & +0.3155 & 0.8666 \\
        \bottomrule
    \end{tabular}
    \vspace{-2mm}
\end{table}

\subsection{\texorpdfstring{\rev{Qualitative Comparison}}{Qualitative Comparison}}
\label{subsec:exp-qualitative}

\begin{figure}[h!]
    \centering
    \includegraphics[width=\linewidth,trim=0 178 0 0,clip]{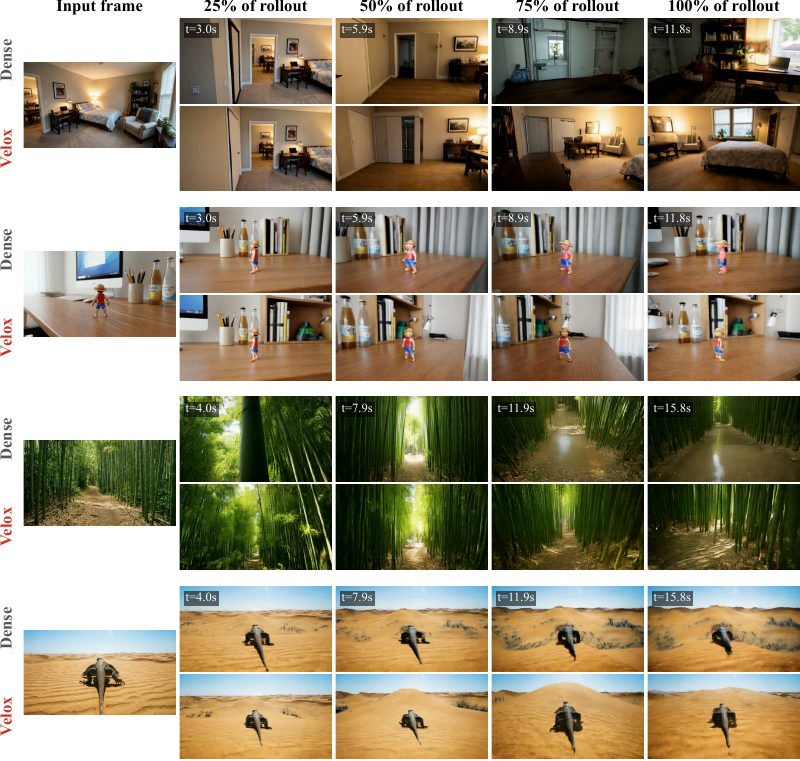}
    \vspace{-1mm}
    \caption{\rev{\textbf{Qualitative comparison with dense attention} at $25\%$ to $100\%$ of the rollout. Late in the rollout, dense attention leaves the bedroom for a dark room and fades the figurine, while \method{} stays consistent with the input. Cases come from the $7$ of $20$ reviewed pairs where \method{} is clearly better (Appendix~\ref{app:more-qualitative}).}}
    \label{fig:qualitative}
    \vspace{-3mm}
\end{figure}

\rev{Aggregate scores do not reveal how a rollout evolves over time. We therefore also compare dense attention and \method{} visually, using a shared input frame and case inputs and sampling frames at fixed fractions of the rollout. Among $20$ reviewed pairs, \method{} stays clearly closer to the input scene in $7$, the two are close in $7$, and the result is mixed in $6$. Figure~\ref{fig:qualitative} shows two of the seven better cases, in which dense attention drifts late in the rollout while \method{} preserves the scene. Appendix~\ref{app:more-qualitative} presents the other five cases, and the project page provides side-by-side videos.}

\section{Conclusion}

We present \method{}, a training-free router that projects spatial memory into the current latent camera view and uses frustum slab overlap to modulate layer-specific block importance while retaining model-derived fallback and local temporal safeguards. At chunk 48, \method{} accelerates attention by $4.36\times$ and denoising by $2.43\times$ with asynchronous geometry construction on one H200. Across all 158 WBench navigation cases, it remains within $0.15$ API-free average points of dense LingBot-Fast-Cam \rev{and stays level with it on the 19-metric overall average}. These results show that region-level spatial correspondence can turn persistent world memory into a practical routing signal for long-horizon generation.

{
    \small

\begin{thebibliography}{94}
\providecommand{\natexlab}[1]{#1}
\providecommand{\url}[1]{\texttt{#1}}
\expandafter\ifx\csname urlstyle\endcsname\relax
  \providecommand{\doi}[1]{doi: #1}\else
  \providecommand{\doi}{doi: \begingroup \urlstyle{rm}\Url}\fi

\bibitem[Alonso et~al.(2024)Alonso, Jelley, Micheli, Kanervisto, Storkey,
  Pearce, and Fleuret]{alonso2024diffusion}
Eloi Alonso, Adam Jelley, Vincent Micheli, Anssi Kanervisto, Amos~J. Storkey,
  Tim Pearce, and Fran{\c{c}}ois Fleuret.
\newblock Diffusion for world modeling: Visual details matter in atari.
\newblock \emph{Advances in Neural Information Processing Systems},
  37:\penalty0 58757--58791, 2024.

\bibitem[Bai et~al.(2025)Bai, Xia, Fu, Wang, Mu, Cao, Liu, Hu, Bai, Wan,
  et~al.]{bai2025recammaster}
Jianhong Bai, Menghan Xia, Xiao Fu, Xintao Wang, Lianrui Mu, Jinwen Cao, Zuozhu
  Liu, Haoji Hu, Xiang Bai, Pengfei Wan, et~al.
\newblock Recammaster: Camera-controlled generative rendering from a single
  video.
\newblock \emph{arXiv preprint arXiv:2503.11647}, 2025.

\bibitem[Beltagy et~al.(2020)Beltagy, Peters, and Cohan]{beltagy2020longformer}
Iz~Beltagy, Matthew~E Peters, and Arman Cohan.
\newblock Longformer: The long-document transformer.
\newblock \emph{arXiv preprint arXiv:2004.05150}, 2020.

\bibitem[Blattmann et~al.(2023)Blattmann, Dockhorn, Kulal, Mendelevitch,
  Kilian, Lorenz, Levi, English, Voleti, Letts, et~al.]{blattmann2023stable}
Andreas Blattmann, Tim Dockhorn, Sumith Kulal, Daniel Mendelevitch, Maciej
  Kilian, Dominik Lorenz, Yam Levi, Zion English, Vikram Voleti, Adam Letts,
  et~al.
\newblock Stable video diffusion: Scaling latent video diffusion models to
  large datasets.
\newblock \emph{arXiv preprint arXiv:2311.15127}, 2023.

\bibitem[Bruce et~al.(2024)Bruce, Dennis, Edwards, Parker-Holder, Shi, Hughes,
  Lai, Mavalankar, Steigerwald, Apps, et~al.]{bruce2024genie}
Jake Bruce, Michael~D Dennis, Ashley Edwards, Jack Parker-Holder, Yuge Shi,
  Edward Hughes, Matthew Lai, Aditi Mavalankar, Richie Steigerwald, Chris Apps,
  et~al.
\newblock Genie: Generative interactive environments.
\newblock In \emph{Forty-first International Conference on Machine Learning},
  2024.

\bibitem[Cabon et~al.(2025)Cabon, Stoffl, Antsfeld, Csurka, Chidlovskii,
  Revaud, and Leroy]{cabon2025must3r}
Yohann Cabon, Lucas Stoffl, Leonid Antsfeld, Gabriela Csurka, Boris
  Chidlovskii, Jerome Revaud, and Vincent Leroy.
\newblock Must3r: Multi-view network for stereo 3d reconstruction.
\newblock In \emph{Proceedings of the Computer Vision and Pattern Recognition
  Conference}, pp.\  1050--1060, 2025.

\bibitem[Che et~al.(2024)Che, He, Liu, Jin, and Chen]{che2024gamegen}
Haoxuan Che, Xuanhua He, Quande Liu, Cheng Jin, and Hao Chen.
\newblock Gamegen-x: Interactive open-world game video generation.
\newblock \emph{arXiv preprint arXiv:2411.00769}, 2024.

\bibitem[Chen et~al.(2024{\natexlab{a}})Chen, Mart{\'i}~Mons{\'o}, Du,
  Simchowitz, Tedrake, and Sitzmann]{chen2024diffusion}
Boyuan Chen, Diego Mart{\'i}~Mons{\'o}, Yilun Du, Max Simchowitz, Russ Tedrake,
  and Vincent Sitzmann.
\newblock Diffusion forcing: Next-token prediction meets full-sequence
  diffusion.
\newblock \emph{Advances in Neural Information Processing Systems},
  37:\penalty0 24081--24125, 2024{\natexlab{a}}.

\bibitem[Chen et~al.(2025{\natexlab{a}})Chen, Lin, Yang, Lin, Zhu, Fan, Zhang,
  Chen, Chen, Ma, et~al.]{chen2025skyreels}
Guibin Chen, Dixuan Lin, Jiangping Yang, Chunze Lin, Juncheng Zhu, Mingyuan
  Fan, Hao Zhang, Sheng Chen, Zheng Chen, Chengchen Ma, et~al.
\newblock Skyreels-v2: Infinite-length film generative model.
\newblock \emph{arXiv preprint arXiv:2504.13074}, 2025{\natexlab{a}}.

\bibitem[Chen et~al.(2024{\natexlab{b}})Chen, Zhang, Cun, Xia, Wang, Weng, and
  Shan]{chen2024videocrafter2}
Haoxin Chen, Yong Zhang, Xiaodong Cun, Menghan Xia, Xintao Wang, Chao Weng, and
  Ying Shan.
\newblock Videocrafter2: Overcoming data limitations for high-quality video
  diffusion models.
\newblock In \emph{Proceedings of the IEEE/CVF Conference on Computer Vision
  and Pattern Recognition}, pp.\  7310--7320, 2024{\natexlab{b}}.

\bibitem[Chen et~al.(2025{\natexlab{b}})Chen, Zhou, Zhao, Wang, Zhang, Huang,
  Sun, Wen, and Li]{chen2025flexworld}
Luxi Chen, Zihan Zhou, Min Zhao, Yikai Wang, Ge~Zhang, Wenhao Huang, Hao Sun,
  Ji-Rong Wen, and Chongxuan Li.
\newblock Flexworld: Progressively expanding 3d scenes for flexible-view
  synthesis.
\newblock \emph{arXiv preprint arXiv:2503.13265}, 2025{\natexlab{b}}.

\bibitem[Dai et~al.(2019)Dai, Yang, Yang, Carbonell, Le, and
  Salakhutdinov]{dai2019transformer}
Zihang Dai, Zhilin Yang, Yiming Yang, Jaime Carbonell, Quoc~V Le, and Ruslan
  Salakhutdinov.
\newblock Transformer-xl: Attentive language models beyond a fixed-length
  context.
\newblock \emph{arXiv preprint arXiv:1901.02860}, 2019.

\bibitem[Dao(2023)]{dao2023flashattention}
Tri Dao.
\newblock Flashattention-2: Faster attention with better parallelism and work
  partitioning.
\newblock \emph{arXiv preprint arXiv:2307.08691}, 2023.

\bibitem[Dao et~al.(2022)Dao, Fu, Ermon, Rudra, and
  R{\'e}]{dao2022flashattention}
Tri Dao, Dan Fu, Stefano Ermon, Atri Rudra, and Christopher R{\'e}.
\newblock Flashattention: Fast and memory-efficient exact attention with
  io-awareness.
\newblock \emph{Advances in Neural Information Processing Systems},
  35:\penalty0 16344--16359, 2022.

\bibitem[Decart et~al.(2024)Decart, McIntyre, Campbell, Chen, and
  Wachen]{oasis2024}
Etched Decart, Quinn McIntyre, Spruce Campbell, Xinlei Chen, and Robert Wachen.
\newblock Oasis: A universe in a transformer.
\newblock \emph{URL: https://oasis-model. github. io}, 2024.

\bibitem[Ding et~al.(2025)Ding, Li, Su, Zhang, Deng, Stoica, and
  Zhang]{ding2025efficientvditefficientvideodiffusion}
Hangliang Ding, Dacheng Li, Runlong Su, Peiyuan Zhang, Zhijie Deng, Ion Stoica,
  and Hao Zhang.
\newblock Efficient-vdit: Efficient video diffusion transformers with attention
  tile, 2025.
\newblock URL \url{https://arxiv.org/abs/2502.06155}.

\bibitem[Engstler et~al.(2024)Engstler, Vedaldi, Laina, and
  Rupprecht]{engstler2024invisible}
Paul Engstler, Andrea Vedaldi, Iro Laina, and Christian Rupprecht.
\newblock Invisible stitch: Generating smooth 3d scenes with depth inpainting.
\newblock In \emph{Arxiv}, 2024.

\bibitem[Esser et~al.(2024)Esser, Kulal, Blattmann, Entezari, M{\"u}ller,
  Saini, Levi, Lorenz, Sauer, Boesel, et~al.]{esser2024scaling}
Patrick Esser, Sumith Kulal, Andreas Blattmann, Rahim Entezari, Jonas
  M{\"u}ller, Harry Saini, Yam Levi, Dominik Lorenz, Axel Sauer, Frederic
  Boesel, et~al.
\newblock Scaling rectified flow transformers for high-resolution image
  synthesis.
\newblock In \emph{Forty-first international conference on machine learning},
  2024.

\bibitem[Feng et~al.(2024)Feng, Zhang, Yang, Xiao, Shu, Liu, Zheng, Huang, Liu,
  and Zhang]{feng2024matrix}
Ruili Feng, Han Zhang, Zhantao Yang, Jie Xiao, Zhilei Shu, Zhiheng Liu, Andy
  Zheng, Yukun Huang, Yu~Liu, and Hongyang Zhang.
\newblock The matrix: Infinite-horizon world generation with real-time moving
  control.
\newblock \emph{arXiv preprint arXiv:2412.03568}, 2024.

\bibitem[Feng et~al.(2026)Feng, Fan, Qin, Yang, Wu, Li, Li, An, Huang, Wang,
  et~al.]{feng2026worldcache}
Weilun Feng, Guoxin Fan, Haotong Qin, Chuanguang Yang, Mingqiang Wu, Yuqi Li,
  Xiangqi Li, Zhulin An, Libo Huang, Dingrui Wang, et~al.
\newblock Worldcache: Accelerating world models for free via heterogeneous
  token caching.
\newblock \emph{arXiv preprint arXiv:2603.06331}, 2026.

\bibitem[Gao et~al.(2025)Gao, Guo, Hoang, Huang, Jiang, Kong, Li, Li, Li, Li,
  et~al.]{gao2025seedance}
Yu~Gao, Haoyuan Guo, Tuyen Hoang, Weilin Huang, Lu~Jiang, Fangyuan Kong, Huixia
  Li, Jiashi Li, Liang Li, Xiaojie Li, et~al.
\newblock Seedance 1.0: Exploring the boundaries of video generation models.
\newblock \emph{arXiv preprint arXiv:2506.09113}, 2025.

\bibitem[Gao et~al.(2026)Gao, Wang, Zeng, Zhu, Cheng, Li, Wang, Xu, Ma, Chen,
  et~al.]{gao2026advancing}
Zelin Gao, Qiuyu Wang, Yanhong Zeng, Jiapeng Zhu, Ka~Leong Cheng, Yixuan Li,
  Hanlin Wang, Yinghao Xu, Shuailei Ma, Yihang Chen, et~al.
\newblock Advancing open-source world models.
\newblock \emph{arXiv preprint arXiv:2601.20540}, 2026.

\bibitem[Gu et~al.(2025)Gu, Mao, and Shou]{gu2025long}
Yuchao Gu, Weijia Mao, and Mike~Zheng Shou.
\newblock Long-context autoregressive video modeling with next-frame
  prediction.
\newblock \emph{arXiv preprint arXiv:2503.19325}, 2025.

\bibitem[Guo et~al.(2024)Guo, Tang, Yang, Zhang, Liu, and
  Han]{guo2024blocksparse}
Junxian Guo, Haotian Tang, Shang Yang, Zhekai Zhang, Zhijian Liu, and Song Han.
\newblock {Block Sparse Attention}.
\newblock \url{https://github.com/mit-han-lab/Block-Sparse-Attention}, 2024.

\bibitem[HaCohen et~al.(2024)HaCohen, Chiprut, Brazowski, Shalem, Moshe,
  Richardson, Levin, Shiran, Zabari, Gordon, et~al.]{hacohen2024ltx}
Yoav HaCohen, Nisan Chiprut, Benny Brazowski, Daniel Shalem, Dudu Moshe, Eitan
  Richardson, Eran Levin, Guy Shiran, Nir Zabari, Ori Gordon, et~al.
\newblock Ltx-video: Realtime video latent diffusion.
\newblock \emph{arXiv preprint arXiv:2501.00103}, 2024.

\bibitem[He et~al.(2024)He, Xu, Guo, Wetzstein, Dai, Li, and
  Yang]{he2024cameractrl}
Hao He, Yinghao Xu, Yuwei Guo, Gordon Wetzstein, Bo~Dai, Hongsheng Li, and
  Ceyuan Yang.
\newblock Cameractrl: Enabling camera control for text-to-video generation.
\newblock \emph{arXiv preprint arXiv:2404.02101}, 2024.

\bibitem[He et~al.(2025)He, Yang, Lin, Xu, Wei, Gui, Zhao, Wetzstein, Jiang,
  and Li]{he2025cameractrl}
Hao He, Ceyuan Yang, Shanchuan Lin, Yinghao Xu, Meng Wei, Liangke Gui, Qi~Zhao,
  Gordon Wetzstein, Lu~Jiang, and Hongsheng Li.
\newblock Cameractrl ii: Dynamic scene exploration via camera-controlled video
  diffusion models.
\newblock \emph{arXiv preprint arXiv:2503.10592}, 2025.

\bibitem[Henschel et~al.(2024)Henschel, Khachatryan, Poghosyan, Hayrapetyan,
  Tadevosyan, Wang, Navasardyan, and Shi]{henschel2024streamingt2v}
Roberto Henschel, Levon Khachatryan, Hayk Poghosyan, Daniil Hayrapetyan, Vahram
  Tadevosyan, Zhangyang Wang, Shant Navasardyan, and Humphrey Shi.
\newblock Streamingt2v: Consistent, dynamic, and extendable long video
  generation from text.
\newblock \emph{arXiv preprint arXiv:2403.14773}, 2024.

\bibitem[Ho et~al.(2020)Ho, Jain, and Abbeel]{ho2020denoising}
Jonathan Ho, Ajay Jain, and Pieter Abbeel.
\newblock Denoising diffusion probabilistic models.
\newblock \emph{NeurIPS}, 33:\penalty0 6840--6851, 2020.

\bibitem[Huang et~al.(2025{\natexlab{a}})Huang, Zheng, Wang, Liu, Wang, Wu,
  Jiang, Li, Lau, Zuo, et~al.]{huang2025voyager}
Tianyu Huang, Wangguandong Zheng, Tengfei Wang, Yuhao Liu, Zhenwei Wang, Junta
  Wu, Jie Jiang, Hui Li, Rynson W.~H. Lau, Wangmeng Zuo, et~al.
\newblock Voyager: Long-range and world-consistent video diffusion for
  explorable 3d scene generation.
\newblock \emph{arXiv preprint arXiv:2506.04225}, 2025{\natexlab{a}}.

\bibitem[Huang et~al.(2025{\natexlab{b}})Huang, Li, He, Zhou, and
  Shechtman]{huang2025self}
Xun Huang, Zhengqi Li, Guande He, Mingyuan Zhou, and Eli Shechtman.
\newblock Self forcing: Bridging the train-test gap in autoregressive video
  diffusion.
\newblock \emph{arXiv preprint arXiv:2506.08009}, 2025{\natexlab{b}}.

\bibitem[HY-World et~al.(2026)HY-World, Cao, Zuo, Wang, Zhang, Wu, Liu, Gong,
  Liu, Yuan, et~al.]{cao2026hyworld}
Team HY-World, Chenjie Cao, Xuhui Zuo, Zhenwei Wang, Yisu Zhang, Junta Wu,
  Zhenyang Liu, Yuning Gong, Yang Liu, Bo~Yuan, et~al.
\newblock Hy-world 2.0: A multi-modal world model for reconstructing,
  generating, and simulating 3d worlds.
\newblock \emph{arXiv preprint arXiv:2604.14268}, 2026.

\bibitem[Jiang et~al.(2024)Jiang, Li, Zhang, Wu, Luo, Ahn, Han, Abdi, Li, Lin,
  et~al.]{jiang2024minference}
Huiqiang Jiang, Yucheng Li, Chengruidong Zhang, Qianhui Wu, Xufang Luo, Surin
  Ahn, Zhenhua Han, Amir Abdi, Dongsheng Li, Chin-Yew Lin, et~al.
\newblock Minference 1.0: Accelerating pre-filling for long-context llms via
  dynamic sparse attention.
\newblock \emph{NeurIPS}, 37:\penalty0 52481--52515, 2024.

\bibitem[Jin et~al.(2024)Jin, Sun, Li, Xu, Jiang, Zhuang, Huang, Song, Mu, and
  Lin]{jin2024pyramidal}
Yang Jin, Zhicheng Sun, Ningyuan Li, Kun Xu, Hao Jiang, Nan Zhuang, Quzhe
  Huang, Yang Song, Yadong Mu, and Zhouchen Lin.
\newblock Pyramidal flow matching for efficient video generative modeling.
\newblock \emph{arXiv preprint arXiv:2410.05954}, 2024.

\bibitem[Keetha et~al.(2025)Keetha, M{\"u}ller, Sch{\"o}nberger, Porzi, Zhang,
  Fischer, Knapitsch, Zauss, Weber, Antunes, et~al.]{keetha2025mapanything}
Nikhil Keetha, Norman M{\"u}ller, Johannes Sch{\"o}nberger, Lorenzo Porzi,
  Yuchen Zhang, Tobias Fischer, Arno Knapitsch, Duncan Zauss, Ethan Weber,
  Nelson Antunes, et~al.
\newblock {MapAnything}: Universal feed-forward metric {3D} reconstruction,
  2025.
\newblock arXiv preprint arXiv:2509.13414.

\bibitem[Kerbl et~al.(2023)Kerbl, Kopanas, Leimk{\"u}hler, and
  Drettakis]{kerbl20233d}
Bernhard Kerbl, Georgios Kopanas, Thomas Leimk{\"u}hler, and George Drettakis.
\newblock 3d gaussian splatting for real-time radiance field rendering.
\newblock \emph{ACM Trans. Graph.}, 42\penalty0 (4):\penalty0 1--14, 2023.

\bibitem[Kim et~al.(2024)Kim, Kang, Choi, and Han]{kim2024fifo}
Jihwan Kim, Junoh Kang, Jinyoung Choi, and Bohyung Han.
\newblock Fifo-diffusion: Generating infinite videos from text without
  training.
\newblock \emph{arXiv preprint arXiv:2405.11473}, 2024.

\bibitem[Kitaev et~al.(2020)Kitaev, Kaiser, and Levskaya]{kitaev2020reformer}
Nikita Kitaev, {\L}ukasz Kaiser, and Anselm Levskaya.
\newblock Reformer: The efficient transformer.
\newblock \emph{ICLR}, 2020.

\bibitem[Kong et~al.(2024)Kong, Tian, Zhang, Min, Dai, Zhou, Xiong, Li, Wu,
  Zhang, et~al.]{kong2024hunyuanvideo}
Weijie Kong, Qi~Tian, Zijian Zhang, Rox Min, Zuozhuo Dai, Jin Zhou, Jiangfeng
  Xiong, Xin Li, Bo~Wu, Jianwei Zhang, et~al.
\newblock Hunyuanvideo: A systematic framework for large video generative
  models.
\newblock \emph{arXiv preprint arXiv:2412.03603}, 2024.

\bibitem[Kwon et~al.(2023)Kwon, Li, Zhuang, Sheng, Zheng, Yu, Gonzalez, Zhang,
  and Stoica]{kwon2023efficient}
Woosuk Kwon, Zhuohan Li, Siyuan Zhuang, Ying Sheng, Lianmin Zheng, Cody~Hao Yu,
  Joseph Gonzalez, Hao Zhang, and Ion Stoica.
\newblock Efficient memory management for large language model serving with
  pagedattention.
\newblock In \emph{Proceedings of the 29th symposium on operating systems
  principles}, pp.\  611--626, 2023.

\bibitem[Lee et~al.(2025)Lee, Park, Suh, and Hwang]{lee2025infinitehip}
Heejun Lee, Geon Park, Jaduk Suh, and Sung~Ju Hwang.
\newblock Infinitehip: Extending language model context up to 3 million tokens
  on a single gpu.
\newblock \emph{arXiv preprint arXiv:2502.08910}, 2025.

\bibitem[Leroy et~al.(2024)Leroy, Cabon, and Revaud]{mast3r_eccv24}
Vincent Leroy, Yohann Cabon, and Jerome Revaud.
\newblock Grounding image matching in 3d with mast3r, 2024.

\bibitem[Li \& Wang(2026)Li and Wang]{li2026jeva}
Qi~Li and Xinchao Wang.
\newblock Jeva: System one models from first principles.
\newblock \emph{arXiv preprint arXiv:2610.18773}, 2026.

\bibitem[Li et~al.(2025)Li, Torr, Vedaldi, and Jakab]{li2025vmem}
Runjia Li, Philip Torr, Andrea Vedaldi, and Tomas Jakab.
\newblock Vmem: Consistent interactive video scene generation with
  surfel-indexed view memory.
\newblock \emph{arXiv preprint arXiv:2506.18903}, 2025.

\bibitem[Li et~al.(2024)Li, Huang, Yang, Venkitesh, Locatelli, Ye, Cai, Lewis,
  and Chen]{li2024snapkv}
Yuhong Li, Yingbing Huang, Bowen Yang, Bharat Venkitesh, Acyr Locatelli,
  Hanchen Ye, Tianle Cai, Patrick Lewis, and Deming Chen.
\newblock Snapkv: Llm knows what you are looking for before generation.
\newblock \emph{Advances in Neural Information Processing Systems},
  37:\penalty0 22947--22970, 2024.

\bibitem[Liu et~al.(2024{\natexlab{a}})Liu, Zhang, Wang, Wei, Qiu, Zhao, Zhang,
  Ye, and Wan]{liu2024timestep}
Feng Liu, Shiwei Zhang, Xiaofeng Wang, Yujie Wei, Haonan Qiu, Yuzhong Zhao,
  Yingya Zhang, Qixiang Ye, and Fang Wan.
\newblock Timestep embedding tells: It's time to cache for video diffusion
  model.
\newblock \emph{arXiv preprint arXiv:2411.19108}, 2024{\natexlab{a}}.

\bibitem[Liu et~al.(2023)Liu, Zaharia, and Abbeel]{liu2023ring}
Hao Liu, Matei Zaharia, and Pieter Abbeel.
\newblock Ring attention with blockwise transformers for near-infinite context.
\newblock \emph{arXiv preprint arXiv:2310.01889}, 2023.

\bibitem[Liu et~al.(2025{\natexlab{a}})Liu, Han, Yan, Wu, Zhu, Wang, Jiang,
  Peng, and Yuan]{liu2025infinitystar}
Jinlai Liu, Jian Han, Bin Yan, Hui Wu, Fengda Zhu, Xing Wang, Yi~Jiang, Bingyue
  Peng, and Zehuan Yuan.
\newblock Infinitystar: Unified spacetime autoregressive modeling for visual
  generation.
\newblock \emph{arXiv preprint arXiv:2511.04675}, 2025{\natexlab{a}}.

\bibitem[Liu et~al.(2024{\natexlab{b}})Liu, Tan, and Wang]{liu2024clear}
Songhua Liu, Zhenxiong Tan, and Xinchao Wang.
\newblock Clear: Conv-like linearization revs pre-trained diffusion
  transformers up.
\newblock \emph{arXiv preprint arXiv:2412.16112}, 2024{\natexlab{b}}.

\bibitem[Liu et~al.(2025{\natexlab{b}})Liu, Yang, Zhang, Zhang, Qiu, You, and
  Yang]{liu2025region}
Ziming Liu, Yifan Yang, Chengruidong Zhang, Yiqi Zhang, Lili Qiu, Yang You, and
  Yuqing Yang.
\newblock Region-adaptive sampling for diffusion transformers.
\newblock \emph{arXiv preprint arXiv:2502.10389}, 2025{\natexlab{b}}.

\bibitem[Ma et~al.(2025)Ma, Huang, Yan, Chen, Duan, Yin, Wan, Ming, Song, Chen,
  et~al.]{ma2025step}
Guoqing Ma, Haoyang Huang, Kun Yan, Liangyu Chen, Nan Duan, Shengming Yin,
  Changyi Wan, Ranchen Ming, Xiaoniu Song, Xing Chen, et~al.
\newblock Step-video-t2v technical report: The practice, challenges, and future
  of video foundation model.
\newblock \emph{arXiv preprint arXiv:2502.10248}, 2025.

\bibitem[Mildenhall et~al.(2021)Mildenhall, Srinivasan, Tancik, Barron,
  Ramamoorthi, and Ng]{mildenhall2021nerf}
Ben Mildenhall, Pratul~P Srinivasan, Matthew Tancik, Jonathan~T Barron, Ravi
  Ramamoorthi, and Ren Ng.
\newblock Nerf: Representing scenes as neural radiance fields for view
  synthesis.
\newblock \emph{Communications of the ACM}, 65\penalty0 (1):\penalty0 99--106,
  2021.

\bibitem[Nawaz et~al.(2026)Nawaz, Heakl, Khan, Shaker, Khan, and
  Khan]{nawaz2026worldcache}
Umair Nawaz, Ahmed Heakl, Ufaq Khan, Abdelrahman Shaker, Salman Khan, and
  Fahad~Shahbaz Khan.
\newblock Worldcache: Content-aware caching for accelerated video world models.
\newblock \emph{arXiv preprint arXiv:2603.22286}, 2026.

\bibitem[Parker-Holder et~al.(2024)Parker-Holder, Ball, Bruce, Dasagi,
  Holsheimer, Kaplanis, Moufarek, Scully, Shar, Shi, Spencer, Yung, Dennis,
  Kenjeyev, Long, Mnih, Chan, Gazeau, Li, Pardo, Wang, Zhang, Besse, Harley,
  Mitenkova, Wang, Clune, Hassabis, Hadsell, Bolton, Singh, and
  Rockt{\"a}schel]{parkerholder2024genie2}
Jack Parker-Holder, Philip Ball, Jake Bruce, Vibhavari Dasagi, Kristian
  Holsheimer, Christos Kaplanis, Alexandre Moufarek, Guy Scully, Jeremy Shar,
  Jimmy Shi, Stephen Spencer, Jessica Yung, Michael Dennis, Sultan Kenjeyev,
  Shangbang Long, Vlad Mnih, Harris Chan, Maxime Gazeau, Bonnie Li, Fabio
  Pardo, Luyu Wang, Lei Zhang, Frederic Besse, Tim Harley, Anna Mitenkova, Jane
  Wang, Jeff Clune, Demis Hassabis, Raia Hadsell, Adrian Bolton, Satinder
  Singh, and Tim Rockt{\"a}schel.
\newblock Genie 2: A large-scale foundation world model, 2024.
\newblock URL
  \url{https://deepmind.google/discover/blog/genie-2-a-large-scale-foundation-world-model/}.

\bibitem[Peebles \& Xie(2023)Peebles and Xie]{peebles2023scalable}
William Peebles and Saining Xie.
\newblock Scalable diffusion models with transformers.
\newblock In \emph{Proceedings of the IEEE/CVF international conference on
  computer vision}, pp.\  4195--4205, 2023.

\bibitem[Ren et~al.(2025)Ren, Shen, Huang, Ling, Lu, Nimier-David, Müller,
  Keller, Fidler, and Gao]{ren2025gen3c}
Xuanchi Ren, Tianchang Shen, Jiahui Huang, Huan Ling, Yifan Lu, Merlin
  Nimier-David, Thomas Müller, Alexander Keller, Sanja Fidler, and Jun Gao.
\newblock Gen3c: 3d-informed world-consistent video generation with precise
  camera control.
\newblock In \emph{Proceedings of the IEEE/CVF Conference on Computer Vision
  and Pattern Recognition}, 2025.

\bibitem[Rombach et~al.(2022)Rombach, Blattmann, Lorenz, Esser, and
  Ommer]{rombach2022high}
Robin Rombach, Andreas Blattmann, Dominik Lorenz, Patrick Esser, and Bj{\"o}rn
  Ommer.
\newblock High-resolution image synthesis with latent diffusion models.
\newblock In \emph{IEEE Conf. Comput. Vis. Pattern Recog.}, 2022.

\bibitem[Sand.ai(2025)]{teng2025magi}
Sand.ai.
\newblock Magi-1: Autoregressive video generation at scale.
\newblock \emph{arXiv preprint arXiv:2505.13211}, 2025.

\bibitem[Shen et~al.(2026)Shen, Zhang, Liao, Li, Tan, Wang, Yan, and
  Wang]{shen2026world}
Qiuhong Shen, Shihua Zhang, Yue Liao, Qi~Li, Zhenxiong Tan, Shizun Wang,
  Shuicheng Yan, and Xinchao Wang.
\newblock World action models: A survey.
\newblock \emph{arXiv preprint arXiv:2606.20781}, 2026.

\bibitem[Team et~al.(2025)Team, Cai, Huang, Kang, Li, Liang, Ma, Ren, Wei, Xie,
  et~al.]{team2025longcat}
Meituan~LongCat Team, Xunliang Cai, Qilong Huang, Zhuoliang Kang, Hongyu Li,
  Shijun Liang, Liya Ma, Siyu Ren, Xiaoming Wei, Rixu Xie, et~al.
\newblock Longcat-video technical report.
\newblock \emph{arXiv preprint arXiv:2510.22200}, 2025.

\bibitem[Valevski et~al.(2024)Valevski, Leviathan, Arar, and
  Fruchter]{valevski2024diffusion}
Dani Valevski, Yaniv Leviathan, Moab Arar, and Shlomi Fruchter.
\newblock Diffusion models are real-time game engines.
\newblock \emph{arXiv preprint arXiv:2408.14837}, 2024.

\bibitem[Wan et~al.(2025)Wan, Wang, Ai, Wen, Mao, Xie, Chen, Yu, Zhao, Yang,
  Zeng, Wang, Zhang, Zhou, Wang, Chen, Zhu, Zhao, Yan, Huang, Feng, Zhang, Li,
  Wu, Chu, Feng, Zhang, Sun, Fang, Wang, Gui, Weng, Shen, Lin, Wang, Wang,
  Zhou, Wang, Shen, Yu, Shi, Huang, Xu, Kou, Lv, Li, Liu, Wang, Zhang, Huang,
  Li, Wu, Liu, Pan, Zheng, Hong, Shi, Feng, Jiang, Han, Wu, and
  Liu]{wan2025wan}
Team Wan, Ang Wang, Baole Ai, Bin Wen, Chaojie Mao, Chen-Wei Xie, Di~Chen,
  Feiwu Yu, Haiming Zhao, Jianxiao Yang, Jianyuan Zeng, Jiayu Wang, Jingfeng
  Zhang, Jingren Zhou, Jinkai Wang, Jixuan Chen, Kai Zhu, Kang Zhao, Keyu Yan,
  Lianghua Huang, Mengyang Feng, Ningyi Zhang, Pandeng Li, Pingyu Wu, Ruihang
  Chu, Ruili Feng, Shiwei Zhang, Siyang Sun, Tao Fang, Tianxing Wang, Tianyi
  Gui, Tingyu Weng, Tong Shen, Wei Lin, Wei Wang, Wei Wang, Wenmeng Zhou, Wente
  Wang, Wenting Shen, Wenyuan Yu, Xianzhong Shi, Xiaoming Huang, Xin Xu, Yan
  Kou, Yangyu Lv, Yifei Li, Yijing Liu, Yiming Wang, Yingya Zhang, Yitong
  Huang, Yong Li, You Wu, Yu~Liu, Yulin Pan, Yun Zheng, Yuntao Hong, Yupeng
  Shi, Yutong Feng, Zeyinzi Jiang, Zhen Han, Zhi-Fan Wu, and Ziyu Liu.
\newblock Wan: Open and advanced large-scale video generative models.
\newblock \emph{arXiv preprint arXiv:2503.20314}, 2025.

\bibitem[Wang et~al.(2024{\natexlab{a}})Wang, Karaev, Rupprecht, and
  Novotny]{wang2024vggsfm}
Jianyuan Wang, Nikita Karaev, Christian Rupprecht, and David Novotny.
\newblock Vggsfm: Visual geometry grounded deep structure from motion.
\newblock In \emph{Proceedings of the IEEE/CVF Conference on Computer Vision
  and Pattern Recognition}, pp.\  21686--21697, 2024{\natexlab{a}}.

\bibitem[Wang et~al.(2025{\natexlab{a}})Wang, Chen, Karaev, Vedaldi, Rupprecht,
  and Novotny]{wang2025vggt}
Jianyuan Wang, Minghao Chen, Nikita Karaev, Andrea Vedaldi, Christian
  Rupprecht, and David Novotny.
\newblock Vggt: Visual geometry grounded transformer.
\newblock In \emph{Proceedings of the IEEE/CVF Conference on Computer Vision
  and Pattern Recognition}, 2025{\natexlab{a}}.

\bibitem[Wang et~al.(2025{\natexlab{b}})Wang, Zhang, Holynski, Efros, and
  Kanazawa]{wang2025continuous}
Qianqian Wang, Yifei Zhang, Aleksander Holynski, Alexei~A Efros, and Angjoo
  Kanazawa.
\newblock Continuous 3d perception model with persistent state.
\newblock \emph{arXiv preprint arXiv:2501.12387}, 2025{\natexlab{b}}.

\bibitem[Wang et~al.(2024{\natexlab{b}})Wang, Leroy, Cabon, Chidlovskii, and
  Revaud]{wang2024dust3r}
Shuzhe Wang, Vincent Leroy, Yohann Cabon, Boris Chidlovskii, and Jerome Revaud.
\newblock Dust3r: Geometric 3d vision made easy.
\newblock In \emph{Proceedings of the IEEE/CVF Conference on Computer Vision
  and Pattern Recognition}, pp.\  20697--20709, 2024{\natexlab{b}}.

\bibitem[Wang et~al.(2020)Wang, Li, Khabsa, Fang, and Ma]{wang2020linformer}
Sinong Wang, Belinda~Z Li, Madian Khabsa, Han Fang, and Hao Ma.
\newblock Linformer: Self-attention with linear complexity.
\newblock \emph{arXiv preprint arXiv:2006.04768}, 2020.

\bibitem[Wang et~al.(2026)Wang, Lin, Yoon, Cho, Zhang, and
  Bansal]{wang2026anchorweave}
Zun Wang, Han Lin, Jaehong Yoon, Jaemin Cho, Yue Zhang, and Mohit Bansal.
\newblock Anchorweave: World-consistent video generation with retrieved local
  spatial memories.
\newblock \emph{arXiv preprint arXiv:2602.14941}, 2026.

\bibitem[Wu et~al.(2025)Wu, Yang, Po, Xu, Liu, Lin, and
  Wetzstein]{wu2025spatialmemory}
Tong Wu, Shuai Yang, Ryan Po, Yinghao Xu, Ziwei Liu, Dahua Lin, and Gordon
  Wetzstein.
\newblock Video world models with long-term spatial memory.
\newblock \emph{arXiv preprint arXiv:2506.05284}, 2025.

\bibitem[Xi et~al.(2025)Xi, Yang, Zhao, Xu, Li, Li, Lin, Cai, Zhang, Li,
  et~al.]{xi2025sparse}
Haocheng Xi, Shuo Yang, Yilong Zhao, Chenfeng Xu, Muyang Li, Xiuyu Li, Yujun
  Lin, Han Cai, Jintao Zhang, Dacheng Li, et~al.
\newblock Sparse videogen: Accelerating video diffusion transformers with
  spatial-temporal sparsity.
\newblock \emph{arXiv preprint arXiv:2502.01776}, 2025.

\bibitem[Xia et~al.(2025)Xia, Ling, Fu, Wang, Li, Xiao, and
  Cui]{xia2025training}
Yifei Xia, Suhan Ling, Fangcheng Fu, Yujie Wang, Huixia Li, Xuefeng Xiao, and
  Bin Cui.
\newblock Training-free and adaptive sparse attention for efficient long video
  generation.
\newblock \emph{arXiv preprint arXiv:2502.21079}, 2025.

\bibitem[Xiao et~al.(2025)Xiao, Lan, Zhou, Ouyang, Yang, Zeng, and
  Pan]{xiao2025worldmem}
Zeqi Xiao, Yushi Lan, Yifan Zhou, Wenqi Ouyang, Shuai Yang, Yanhong Zeng, and
  Xingang Pan.
\newblock Worldmem: Long-term consistent world simulation with memory.
\newblock \emph{arXiv preprint arXiv:2504.12369}, 2025.

\bibitem[Xie et~al.(2024)Xie, Xu, Hong, Tan, Liu, Liu, Kaufman, and
  Zhou]{xie2024progressive}
Desai Xie, Zhan Xu, Yicong Hong, Hao Tan, Difan Liu, Feng Liu, Arie Kaufman,
  and Yang Zhou.
\newblock Progressive autoregressive video diffusion models.
\newblock \emph{arXiv preprint arXiv:2410.08151}, 2024.

\bibitem[Yang et~al.(2025{\natexlab{a}})Yang, Sax, Liang, Henaff, Tang, Cao,
  Chai, Meier, and Feiszli]{Yang_2025_Fast3R}
Jianing Yang, Alexander Sax, Kevin~J. Liang, Mikael Henaff, Hao Tang, Ang Cao,
  Joyce Chai, Franziska Meier, and Matt Feiszli.
\newblock Fast3r: Towards 3d reconstruction of 1000+ images in one forward
  pass.
\newblock In \emph{Proceedings of the IEEE/CVF Conference on Computer Vision
  and Pattern Recognition (CVPR)}, June 2025{\natexlab{a}}.

\bibitem[Yang et~al.(2024{\natexlab{a}})Yang, Li, Fang, Chen, Yu, Fu, Yang, and
  Ye]{yang2024playable}
Mingyu Yang, Junyou Li, Zhongbin Fang, Sheng Chen, Yangbin Yu, Qiang Fu, Wei
  Yang, and Deheng Ye.
\newblock Playable game generation.
\newblock \emph{arXiv preprint arXiv:2412.00887}, 2024{\natexlab{a}}.

\bibitem[Yang et~al.(2025{\natexlab{b}})Yang, Xi, Zhao, Li, Zhang, Cai, Lin,
  Li, Xu, Peng, et~al.]{yang2025sparsevideogen2}
Shuo Yang, Haocheng Xi, Yilong Zhao, Muyang Li, Jintao Zhang, Han Cai, Yujun
  Lin, Xiuyu Li, Chenfeng Xu, Kelly Peng, et~al.
\newblock Sparse videogen2: Accelerate video generation with sparse attention
  via semantic-aware permutation.
\newblock \emph{arXiv preprint arXiv:2505.18875}, 2025{\natexlab{b}}.

\bibitem[Yang et~al.(2025{\natexlab{c}})Yang, Xu, Luan, Zhan, Qiu, Shi, Li,
  Yang, Zhang, Yu, et~al.]{yang2025omnicam}
Xiaoda Yang, Jiayang Xu, Kaixuan Luan, Xinyu Zhan, Hongshun Qiu, Shijun Shi,
  Hao Li, Shuai Yang, Li~Zhang, Checheng Yu, et~al.
\newblock Omnicam: Unified multimodal video generation via camera control.
\newblock \emph{arXiv preprint arXiv:2504.02312}, 2025{\natexlab{c}}.

\bibitem[Yang et~al.(2024{\natexlab{b}})Yang, Teng, Zheng, Ding, Huang, Xu,
  Yang, Hong, Zhang, Feng, et~al.]{yang2024cogvideox}
Zhuoyi Yang, Jiayan Teng, Wendi Zheng, Ming Ding, Shiyu Huang, Jiazheng Xu,
  Yuanming Yang, Wenyi Hong, Xiaohan Zhang, Guanyu Feng, et~al.
\newblock Cogvideox: Text-to-video diffusion models with an expert transformer.
\newblock \emph{arXiv preprint arXiv:2408.06072}, 2024{\natexlab{b}}.

\bibitem[Ying et~al.(2026)Ying, Hu, Ren, Li, Chen, Wang, Cao, Cai, and
  Ding]{ying2026wbench}
Kaining Ying, Hengrui Hu, Siyu Ren, Jiamu Li, Fengjiao Chen, Ziwen Wang, Xuezhi
  Cao, Xunliang Cai, and Henghui Ding.
\newblock Wbench: A comprehensive multi-turn benchmark for interactive video
  world model evaluation.
\newblock \emph{arXiv preprint arXiv:2605.25874}, 2026.

\bibitem[Yu et~al.(2024{\natexlab{a}})Yu, Duan, Herrmann, Freeman, and
  Wu]{yu2024wonderworld}
Hong-Xing Yu, Haoyi Duan, Charles Herrmann, William~T. Freeman, and Jiajun Wu.
\newblock Wonderworld: Interactive 3d scene generation from a single image.
\newblock \emph{arXiv:2406.09394}, 2024{\natexlab{a}}.

\bibitem[Yu et~al.(2024{\natexlab{b}})Yu, Duan, Hur, Sargent, Rubinstein,
  Freeman, Cole, Sun, Snavely, Wu, et~al.]{yu2024wonderjourney}
Hong-Xing Yu, Haoyi Duan, Junhwa Hur, Kyle Sargent, Michael Rubinstein,
  William~T Freeman, Forrester Cole, Deqing Sun, Noah Snavely, Jiajun Wu,
  et~al.
\newblock Wonderjourney: Going from anywhere to everywhere.
\newblock In \emph{Proceedings of the IEEE/CVF Conference on Computer Vision
  and Pattern Recognition}, pp.\  6658--6667, 2024{\natexlab{b}}.

\bibitem[Yu et~al.(2025{\natexlab{a}})Yu, Bai, Qin, Liu, Wang, Wan, Zhang, and
  Liu]{yu2025context}
Jiwen Yu, Jianhong Bai, Yiran Qin, Quande Liu, Xintao Wang, Pengfei Wan,
  Di~Zhang, and Xihui Liu.
\newblock Context as memory: Scene-consistent interactive long video generation
  with memory retrieval.
\newblock \emph{arXiv preprint arXiv:2506.03141}, 2025{\natexlab{a}}.

\bibitem[Yu et~al.(2025{\natexlab{b}})Yu, Qin, Wang, Wan, Zhang, and
  Liu]{yu2025gamefactory}
Jiwen Yu, Yiran Qin, Xintao Wang, Pengfei Wan, Di~Zhang, and Xihui Liu.
\newblock Gamefactory: Creating new games with generative interactive videos.
\newblock \emph{arXiv preprint arXiv:2501.08325}, 2025{\natexlab{b}}.

\bibitem[YU et~al.(2025)YU, Hu, Xing, and Shan]{yu2025trajectorycrafter}
Mark YU, Wenbo Hu, Jinbo Xing, and Ying Shan.
\newblock Trajectorycrafter: Redirecting camera trajectory for monocular videos
  via diffusion models.
\newblock \emph{arXiv preprint arXiv:2503.05638}, 2025.

\bibitem[Yu et~al.(2026)Yu, Tan, Chen, Liu, and Wang]{yu2026vifeedit}
Ruonan Yu, Zhenxiong Tan, Zigeng Chen, Songhua Liu, and Xinchao Wang.
\newblock Vifeedit: A video-free tuner of your video diffusion transformer.
\newblock \emph{arXiv preprint arXiv:2603.15478}, 2026.

\bibitem[Yu et~al.(2024{\natexlab{c}})Yu, Xing, Yuan, Hu, Li, Huang, Gao, Wong,
  Shan, and Tian]{yu2024viewcrafter}
Wangbo Yu, Jinbo Xing, Li~Yuan, Wenbo Hu, Xiaoyu Li, Zhipeng Huang, Xiangjun
  Gao, Tien-Tsin Wong, Ying Shan, and Yonghong Tian.
\newblock Viewcrafter: Taming video diffusion models for high-fidelity novel
  view synthesis.
\newblock \emph{arXiv preprint arXiv:2409.02048}, 2024{\natexlab{c}}.

\bibitem[Zaheer et~al.(2020)Zaheer, Guruganesh, Dubey, Ainslie, Alberti,
  Ontanon, Pham, Ravula, Wang, Yang, et~al.]{zaheer2020big}
Manzil Zaheer, Guru Guruganesh, Kumar~Avinava Dubey, Joshua Ainslie, Chris
  Alberti, Santiago Ontanon, Philip Pham, Anirudh Ravula, Qifan Wang, Li~Yang,
  et~al.
\newblock Big bird: Transformers for longer sequences.
\newblock \emph{Advances in neural information processing systems},
  33:\penalty0 17283--17297, 2020.

\bibitem[Zhang et~al.(2025{\natexlab{a}})Zhang, Chen, Wang, Liu, Wang, and
  Qiao]{zhang2025accvideo}
Haiyu Zhang, Xinyuan Chen, Yaohui Wang, Xihui Liu, Yunhong Wang, and Yu~Qiao.
\newblock Accvideo: Accelerating video diffusion model with synthetic dataset.
\newblock \emph{arXiv preprint arXiv:2503.19462}, 2025{\natexlab{a}}.

\bibitem[Zhang et~al.(2024)Zhang, Huang, Zhang, Zhu, Chen,
  et~al.]{zhang2024sageattention}
Jintao Zhang, Haofeng Huang, Pengle Zhang, Jun Zhu, Jianfei Chen, et~al.
\newblock Sageattention: Accurate 8-bit attention for plug-and-play inference
  acceleration.
\newblock \emph{arXiv preprint arXiv:2410.02367}, 2024.

\bibitem[Zhang et~al.(2025{\natexlab{b}})Zhang, Xiang, Huang, Wei, Xi, Zhu, and
  Chen]{zhang2025spargeattn}
Jintao Zhang, Chendong Xiang, Haofeng Huang, Jia Wei, Haocheng Xi, Jun Zhu, and
  Jianfei Chen.
\newblock Spargeattention: Accurate and training-free sparse attention
  accelerating any model inference.
\newblock \emph{Proceedings of the 42nd International Conference on Machine
  Learning}, 2025{\natexlab{b}}.

\bibitem[Zhang et~al.(2023{\natexlab{a}})Zhang, Rao, and
  Agrawala]{zhang2023adding}
Lvmin Zhang, Anyi Rao, and Maneesh Agrawala.
\newblock Adding conditional control to text-to-image diffusion models.
\newblock In \emph{Proceedings of the IEEE/CVF international conference on
  computer vision}, pp.\  3836--3847, 2023{\natexlab{a}}.

\bibitem[Zhang et~al.(2025{\natexlab{c}})Zhang, Chen, Su, Ding, Stoica, Liu,
  and Zhang]{zhang2025fast}
Peiyuan Zhang, Yongqi Chen, Runlong Su, Hangliang Ding, Ion Stoica, Zhenghong
  Liu, and Hao Zhang.
\newblock Fast video generation with sliding tile attention.
\newblock \emph{arXiv preprint arXiv:2502.04507}, 2025{\natexlab{c}}.

\bibitem[Zhang et~al.(2025{\natexlab{d}})Zhang, Peng, Wang, Wang, Zhu, Gao, Li,
  Liu, and Zhou]{zhang2025matrixgame}
Yifan Zhang, Chunli Peng, Boyang Wang, Puyi Wang, Qingcheng Zhu, Zedong Gao,
  Eric Li, Yang Liu, and Yahui Zhou.
\newblock Matrix-game: Interactive world foundation model.
\newblock \emph{arXiv}, 2025{\natexlab{d}}.

\bibitem[Zhang et~al.(2023{\natexlab{b}})Zhang, Sheng, Zhou, Chen, Zheng, Cai,
  Song, Tian, R{\'e}, Barrett, et~al.]{zhang2023h2o}
Zhenyu Zhang, Ying Sheng, Tianyi Zhou, Tianlong Chen, Lianmin Zheng, Ruisi Cai,
  Zhao Song, Yuandong Tian, Christopher R{\'e}, Clark Barrett, et~al.
\newblock H2o: Heavy-hitter oracle for efficient generative inference of large
  language models.
\newblock \emph{Advances in Neural Information Processing Systems},
  36:\penalty0 34661--34710, 2023{\natexlab{b}}.

\end{thebibliography}

}

\clearpage
\appendix

\begin{appendices}

\section{Additional Experimental Evidence}
\label{app:evidence}

\subsection{Soft-Fusion Calibration}
\label{app:soft-calibration}

We calibrate the fixed fusion strength on pooled query-key distributions and frustum slab overlap matrices from two navigation trajectories. The calibration contains 48 grids spanning four representative layers, two denoising steps, and three rollout lengths. Positive overlap values have median $0.211$, which motivates rescaling rather than multiplying by raw IoU. Among the tested settings, $\lambda=2$ and $g_{\mathrm{ref}}=0.2$ are the weakest pair that recovers at least $85\%$ of positive-overlap blocks in the fused top-80 while changing at most $25\%$ of the model-only selections. Concretely, positive-overlap recall increases from $0.602$ to $0.852$, and the selected set changes by $9.0\%$ on average. Values at or below $\tau=0.1$ receive no boost.

\subsection{\texorpdfstring{\rev{Additional Qualitative Comparisons}}{Additional Qualitative Comparisons}}
\label{app:more-qualitative}

\begin{figure}[h!]
    \centering
    \includegraphics[width=\linewidth]{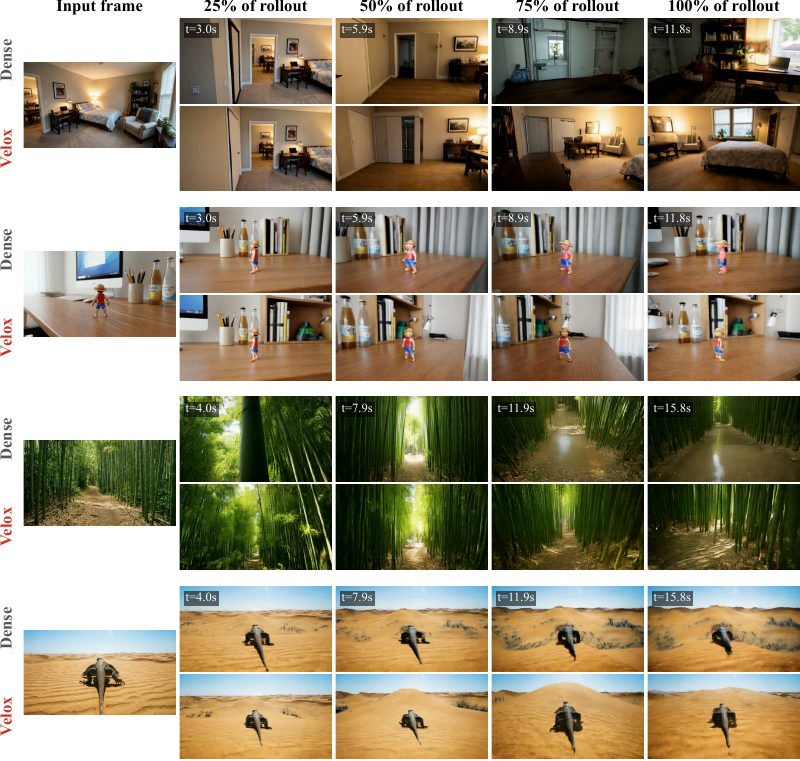}
    \caption{\rev{\textbf{Qualitative comparison with dense attention, four cases.} Shared input frame, then dense LingBot-Fast-Cam (top row) and \method{} (bottom row) at $25\%$, $50\%$, $75\%$, and $100\%$ of the rollout. Timestamps are in seconds. Late in the rollout, dense attention replaces the bedroom with an unrelated dark room and fades the figurine to pale pink. It also turns the bamboo path into a puddle and spreads dark artifacts over the dunes. \method{} stays consistent with the input scene.}}
    \label{fig:qualitative-full}
\end{figure}

\begin{figure}[h!]
    \centering
    \includegraphics[width=\linewidth]{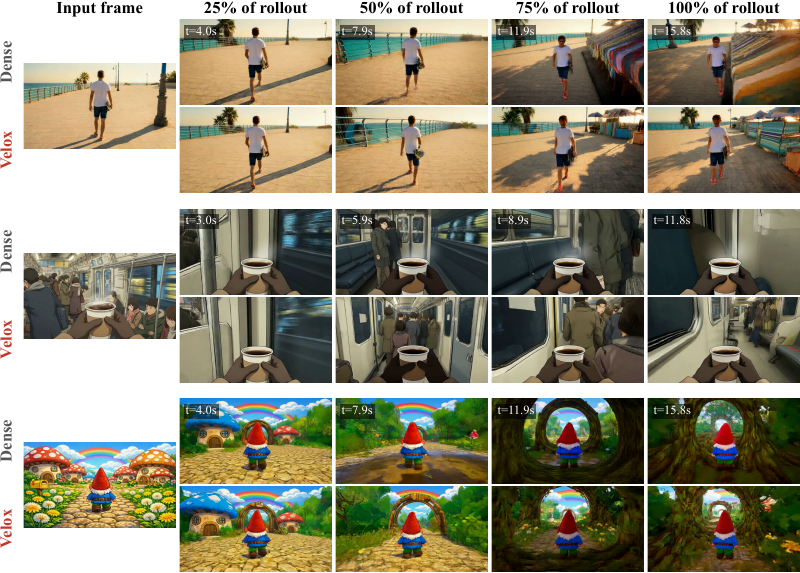}
    \caption{\rev{\textbf{Additional qualitative comparisons.} Same protocol as Figure~\ref{fig:qualitative-full}. In the beach case, dense attention smears texture and distorts the subject. In the subway case, it loses the crowd and ends on a blurred surface. In the gnome case, it turns the cobblestone path into a puddle. \method{} stays closer to the input frame in all three cases.}}
    \label{fig:qualitative-more}
\end{figure}

\rev{We reviewed $20$ paired rollouts of dense LingBot-Fast-Cam and \method{} on WBench cases, which we did not select by their outcome. Each pair shares the input frame and the case inputs, and we sample frames at $25\%$, $50\%$, $75\%$, and $100\%$ of the rollout instead of hand-picking moments. We judged each pair by watching the videos, using three simple scores against the input frame as a screen: matched SIFT features, color histogram correlation, and retained sharpness. Under this protocol, \method{} stays clearly closer to the input scene in $7$ pairs and the two models are close in $7$. The remaining $6$ pairs are mixed, with \method{} ahead at some timestamps and behind at others.

Figures~\ref{fig:qualitative-full} and~\ref{fig:qualitative-more} show all seven cases where \method{} is better. In the bedroom case, dense attention ends in an unrelated dark room, while \method{} returns to the same bedroom with the same bed cover, desk, and armchair. In the figurine case, dense attention fades the red outfit to pale pink. In the bamboo case, it turns the path into a puddle and then a featureless surface, and in the desert case it spreads dark artifacts over the dunes. In the beach case, it smears the towel stalls and distorts the face of the walking person. In the subway case, it loses the crowd and ends on a blurred surface. In the gnome case, both models finally enter a tunnel of roots, but dense attention first turns the cobblestone path into a puddle. \method{} preserves these details in all seven cases. Although these cases agree with the higher spatial consistency of \method{} in Table~\ref{tab:wbench-navi} ($74.3$ versus $73.0$), they are selected, and the paired interval for spatial consistency includes zero (Table~\ref{tab:paired}). We thus present them as illustrations of a failure mode of dense attention, and leave the measurement of average quality to the tables.}

\subsection{\texorpdfstring{\rev{Paired Per-Case Quality}}{Paired Per-Case Quality}}
\label{app:paired}

{\revblue{}Table~\ref{tab:wbench-navi} reports means over the 158 cases, and such means can hide a systematic loss. To examine this, we pair the per-case scores of dense LingBot-Fast-Cam and \method{} from the same runs as Table~\ref{tab:wbench-navi}. For every metric, we compute the per-case difference (\method{} minus dense) and a paired bootstrap $95\%$ interval with $10{,}000$ resamples over cases, using only the cases that the metric scores. Since the API-free and overall averages of Table~\ref{tab:wbench-navi} are coverage-aware aggregates without a per-case value, we test the metrics individually. As a summary, we additionally test the plain mean of eight automatic metrics that score every case, namely aesthetic and imaging quality, background consistency, motion smoothness, temporal flickering, navigation trajectory, segment continuity, and reconstruction consistency. Three metrics with fewer than ten cases are omitted from Table~\ref{tab:paired}.

Six of the seventeen intervals exclude zero. \method{} gains on imaging and aesthetic quality and loses on perspective consistency, subject consistency, motion smoothness, and temporal flickering. Among these losses, only perspective consistency exceeds one point ($-4.7$, $[-8.6,-0.9]$), making it the one material deficit, whereas the eight-metric mean shows no clear difference ($+0.34$, $[-0.38,+1.08]$). Although Block-Sparse lacks per-case reports, its perspective score in Table~\ref{tab:wbench-navi} ($75.7$) lies below that of \method{} ($77.0$), which is consistent with a loss introduced by sparse routing and partly recovered by geometry.

The ten cases with the largest drop in the eight-metric mean lose between $3.8$ and $17.2$ points. They are spread over the three camera-drift groups of Appendix~\ref{app:drift} (four high, three middle, and three low), so the largest losses do not concentrate where the camera drifts most. Interestingly, one of the ten is the bamboo case of Figure~\ref{fig:qualitative-full}, in which dense attention visibly loses the path. This case shows that automatic metrics and visual inspection can disagree, and it motivates reporting both.}

\begin{table}[h!]
    \centering
    \scriptsize
    \caption{\rev{\textbf{Paired per-case comparison of \method{} and dense attention on WBench navigation.} Scores in $[0,100]$. $n$ is the number of cases the metric scores. $\Delta$ is the mean of \method{} minus dense, with a paired bootstrap $95\%$ interval. Bold marks intervals that exclude zero.}}
    \label{tab:paired}
    \setlength{\tabcolsep}{5pt}
    \revblue
    \begin{tabular}{lrrrl}
        \toprule
        Metric & $n$ & Dense & \method{} & $\Delta$ [95\% interval] \\
        \midrule
        Imaging quality & 158 & 61.29 & 62.57 & $\mathbf{+1.28}$ [$+0.84$, $+1.74$] \\
        Aesthetic quality & 158 & 60.97 & 61.51 & $\mathbf{+0.55}$ [$+0.06$, $+1.05$] \\
        Perspective consistency & 103 & 81.74 & 77.04 & $\mathbf{-4.70}$ [$-8.58$, $-0.94$] \\
        Subject consistency & 103 & 87.18 & 86.37 & $\mathbf{-0.81}$ [$-1.58$, $-0.07$] \\
        Temporal flickering & 158 & 92.69 & 92.41 & $\mathbf{-0.28}$ [$-0.45$, $-0.10$] \\
        Motion smoothness & 158 & 96.20 & 95.97 & $\mathbf{-0.23}$ [$-0.35$, $-0.11$] \\
        \midrule
        Spatial consistency & 60 & 73.02 & 74.33 & $+1.31$ [$-0.18$, $+2.83$] \\
        Subject adherence & 74 & 87.97 & 90.41 & $+2.43$ [$-2.16$, $+7.03$] \\
        Segment continuity & 158 & 90.51 & 92.41 & $+1.90$ [$-3.16$, $+6.96$] \\
        Dynamic degree & 158 & 99.37 & 100.00 & $+0.63$ [$+0.00$, $+1.90$] \\
        HPSv3 quality & 158 & 55.77 & 55.94 & $+0.17$ [$-0.82$, $+1.17$] \\
        Background consistency & 158 & 87.94 & 88.03 & $+0.09$ [$-0.22$, $+0.41$] \\
        Causal fidelity & 50 & 21.81 & 21.91 & $+0.09$ [$-3.58$, $+3.68$] \\
        Reconstruction consistency & 158 & 80.45 & 80.50 & $+0.04$ [$-1.66$, $+1.79$] \\
        Visual plausibility & 158 & 56.08 & 55.79 & $-0.29$ [$-0.71$, $+0.14$] \\
        Navigation trajectory & 158 & 79.45 & 78.77 & $-0.68$ [$-2.15$, $+0.85$] \\
        Scene adherence & 158 & 71.84 & 70.19 & $-1.65$ [$-4.30$, $+1.01$] \\
        \midrule
        Mean of the eight automatic metrics & 158 & 81.19 & 81.52 & $+0.34$ [$-0.38$, $+1.08$] \\
        \bottomrule
    \end{tabular}
\end{table}

\subsection{\texorpdfstring{\rev{Robustness to Camera-Following Errors}}{Robustness to Camera-Following Errors}}
\label{app:drift}

{\revblue{}Appendix~\ref{subsec:exp-failures} discusses how a generated chunk that departs from its commanded camera can match a well-covered region to the wrong history. We examine this concern from two angles.

\noindent\textbf{Drift in real rollouts.} WBench measures how closely each generated video follows its commanded camera path through the navigation-trajectory metric, where a lower score indicates more drift. We average this score over both models for each case and split the 158 cases into three groups of near-equal size, fixing the groups before computing any difference between the models. Within each group, we then repeat the paired test of Appendix~\ref{app:paired}. Should wrong matches under drift harm \method{}, the loss would be largest in the high-drift group. Table~\ref{tab:drift} shows no such trend. The eight-metric mean shows no clear difference in any group, the perspective loss is clear only in the middle group, and the only clear navigation loss appears in the low-drift group. In other words, the perspective deficit acts as a roughly uniform offset across groups and does not grow with drift.

\begin{table}[h!]
    \centering
    \scriptsize
    \caption{\rev{\textbf{\method{} minus dense attention, grouped by camera drift.} Drift groups use the navigation-trajectory score averaged over both models, where lower means more drift. Entries are paired mean differences in points with bootstrap $95\%$ intervals. Bold marks intervals that exclude zero. $n$ for perspective consistency is the number of scored cases in the group.}}
    \label{tab:drift}
    \setlength{\tabcolsep}{4pt}
    \revblue
    \begin{tabular}{lcccc}
        \toprule
        Drift group & Cases & Eight-metric mean & Perspective consistency & Navigation trajectory \\
        \midrule
        High (score $0.465$ to $0.751$) & 52 & $+0.41$ [$-0.94$, $+1.81$] & $-2.84$ [$-9.73$, $+3.92$], $n{=}42$ & $+0.62$ [$-2.79$, $+4.17$] \\
        Middle ($0.755$ to $0.854$) & 52 & $+0.42$ [$-1.01$, $+1.93$] & $\mathbf{-7.24}$ [$-14.02$, $-1.86$], $n{=}34$ & $-0.94$ [$-3.22$, $+1.34$] \\
        Low ($0.855$ to $0.987$) & 54 & $+0.18$ [$-0.74$, $+1.15$] & $-4.39$ [$-10.57$, $+1.88$], $n{=}27$ & $\mathbf{-1.68}$ [$-3.18$, $-0.07$] \\
        \bottomrule
    \end{tabular}
\end{table}

\noindent\textbf{A deliberate camera offset.} Since real drift cannot be set by hand, we also induce the mismatch deliberately. Specifically, we rotate the target cameras used by the geometry by a fixed $15^\circ$ yaw while the generator still receives the original commanded cameras. A small sideways shift is added as well, but it is negligible at the camera step sizes of these cases, leaving the yaw as the effective change. As a result, the geometry describes a view that the generator never produces, which mirrors the mismatch discussed above. We run three cases, two of which return to their starting view, on shortened rollouts of about ten chunks that keep the full camera path. Table~\ref{tab:offset} reports the fraction of current blocks that pass the coverage test ($\eta_{\mathrm{cov}}=0.9$) and thus receive the geometric boost. Under the offset, this fraction drops in two of the three cases, and the rows that fail the test fall back to the model-only route. In these cases, \method{} narrows the influence of geometry once the warped views stop agreeing with the stored slabs, instead of trusting the wrong overlap. In the third case, coverage stays high and fusion runs on the shifted geometry. Across all settings, the frames look nearly identical, as expected, because the offset alters only the route and leaves the camera followed by the generator unchanged. This test therefore characterizes how the route responds to a camera mismatch.}

\begin{table}[h!]
    \centering
    \scriptsize
    \caption{\rev{\textbf{Coverage under a deliberate $15^\circ$ geometry-only camera offset.} Fraction of current blocks that pass the coverage test, averaged over the chunks that use \method{}. The generator receives the original cameras in both settings.}}
    \label{tab:offset}
    \setlength{\tabcolsep}{6pt}
    \revblue
    \begin{tabular}{lcc}
        \toprule
        Case & No offset & $15^\circ$ offset \\
        \midrule
        Revisit case A & 0.91 & 0.60 \\
        Case B & 0.39 & 0.29 \\
        Revisit case C (bedroom) & 0.83 & 0.88 \\
        \bottomrule
    \end{tabular}
\end{table}

\subsection{\texorpdfstring{\rev{Model-Only Routing and Selected History Blocks}}{Model-Only Routing and Selected History Blocks}}
\label{app:blocks}

\begin{figure}[h!]
    \centering
    \includegraphics[width=\linewidth]{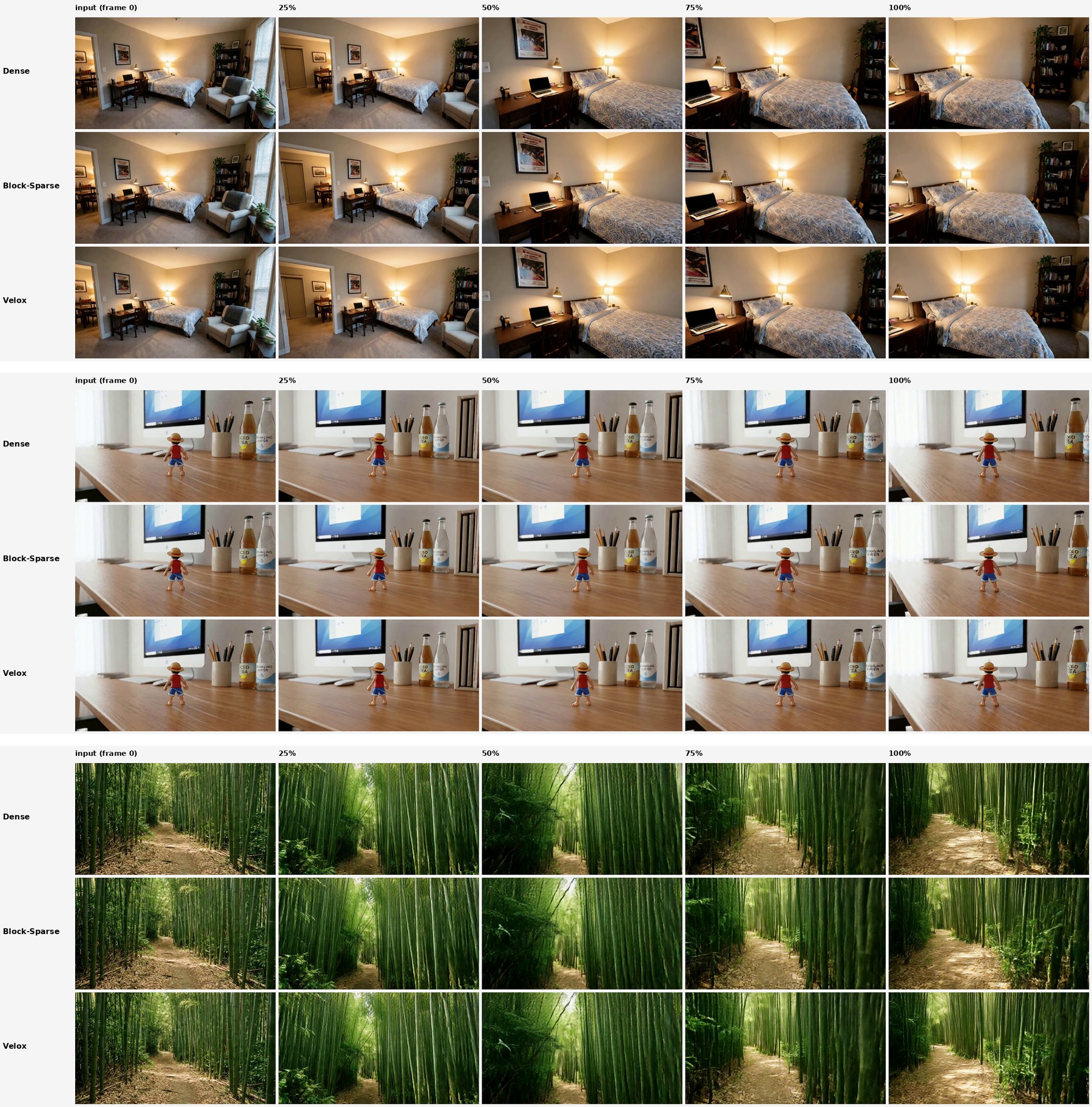}
    \caption{\rev{\textbf{Dense attention, Block-Sparse, and \method{} on shortened rollouts.} Bedroom, figurine, and bamboo cases. Columns show the shared input frame and $25\%$ to $100\%$ of the rollout. On these short rollouts the three models stay close to each other. The long-horizon drift of dense attention appears in Figures~\ref{fig:qualitative-full} and~\ref{fig:qualitative-more}.}}
    \label{fig:blocksparse-short}
\end{figure}

{\revblue\noindent\textbf{Block-Sparse next to dense attention and \method{}.} Figure~\ref{fig:blocksparse-short} adds the model-only route (Block-Sparse) to the qualitative comparison for three of our qualitative cases, with a shared input frame and camera path. These rollouts are shortened to about 80 frames along the same path. At this short length, all three models stay close to the input scene, and the differences emerge late in long rollouts, as Figures~\ref{fig:qualitative-full} and~\ref{fig:qualitative-more} show for dense attention and \method{}.}

\begin{figure}[h!]
    \centering
    \includegraphics[width=\linewidth]{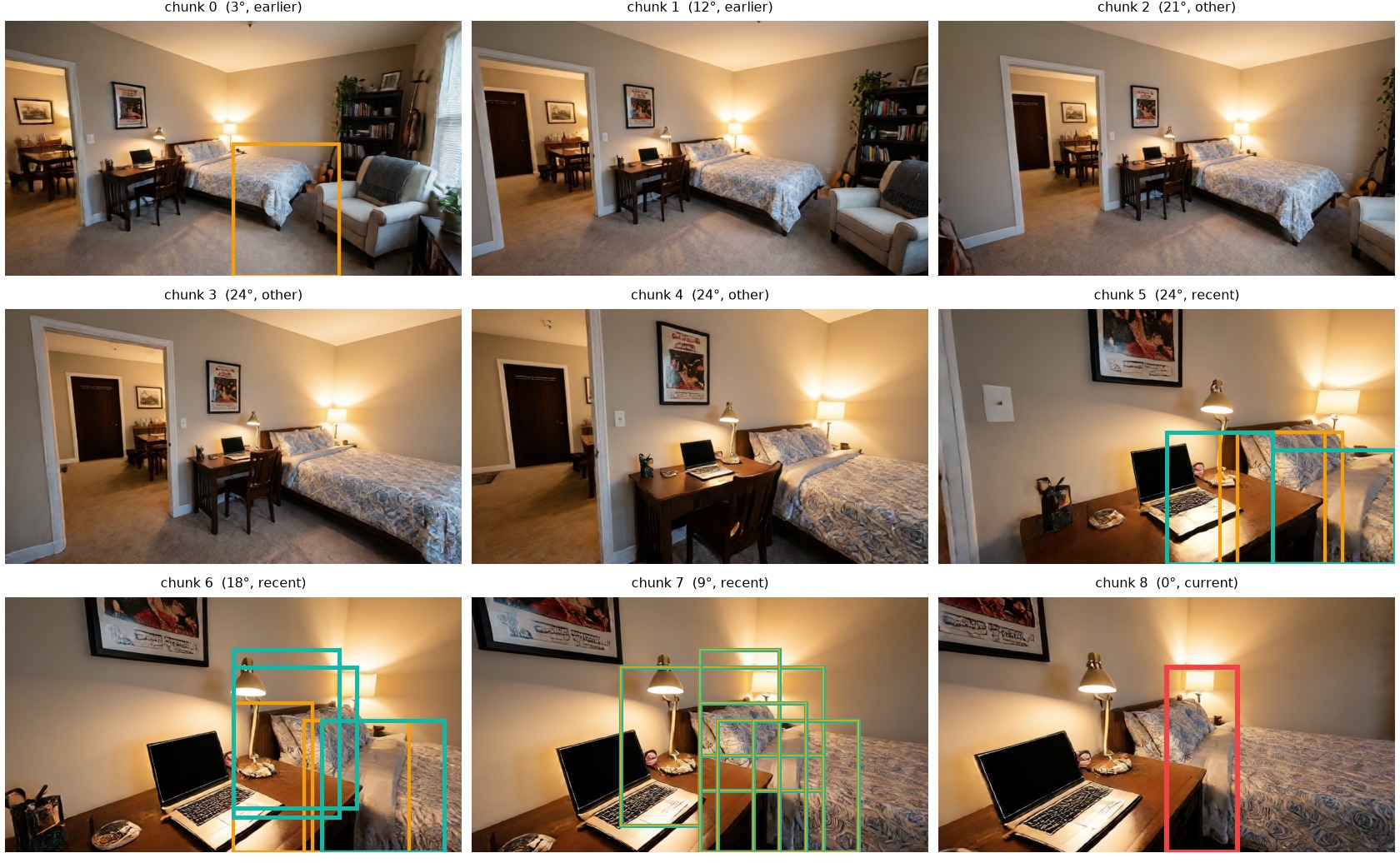}
    \caption{\rev{\textbf{History blocks selected for one query block.} Bedroom case, chunk 8, layer 14, first denoising step. Red: the query block in the current chunk. Amber: top 12 history blocks by model importance alone. Teal: top 12 after soft fusion with slab overlap. Boxes chosen by both routes show both colors. Titles give each chunk's camera angle to chunk 8. Fusion moves the selection onto blocks that overlap the query region.}}
    \label{fig:selected-blocks}
\end{figure}

{\revblue\noindent\textbf{Which history blocks \method{} selects.} Figure~\ref{fig:selected-blocks} visualizes the route of one query block in the bedroom case. We take chunk 8, which faces the same direction as chunk 0 ($3^\circ$ apart) from a closer position, and select the well-covered bed block nearest the image center. We then compare the top 12 history blocks ranked by model importance alone with the top 12 after soft fusion, at layer 14 and the first denoising step. Fusion shifts the selection toward history blocks whose slabs overlap the query block: the mean slab overlap with the query block rises from $0.117$ to $0.201$, and the number of selected blocks with positive overlap grows from 5 to 10 of 12 (from 7 to 15 with 30 blocks). Because the bed stays in view along this short path, its closest matching views lie in the recent chunks 5 to 7, and fusion selects exactly those views. The figure thus confirms that geometry steers the route toward the region that the query block sees. Finally, this run splits the attention heads across two GPUs, so the model importance in the figure is pooled over the 20 heads on one GPU, whereas the slab overlap is complete.}

\clearpage
\section{Experimental Implementation Details}
\label{app:impl-settings}

\method{} is instantiated on the LingBot-Fast-Cam inference stack~\citep{gao2026advancing} with a Wan-style video diffusion transformer. Its routing backend uses frustum slab matching: stored chunks provide source-camera token rectangles and robust depth intervals, target query blocks obtain Q-depth through multiwarp, and Eq.~\eqref{eq:frustum-iou} provides the world-space OBB overlap score.

\subsection{Reported Routing Configuration}
\label{app:routing-config}

The reported configuration applies the fixed fusion in Eq.~\eqref{eq:soft-geo-fusion} only when $\rho_i\geq0.9$. It uses $\lambda=2$, $g_{\mathrm{ref}}=0.2$, $\tau=0.1$, and $K_{\mathrm{cov}}=80$. Rows below the coverage threshold retain the model-only cumulative-mass route and are capped at $K_{\mathrm{unc}}=150$ without a minimum route-size floor. The previous chunk is reserved inside these budgets. Global Gilbert neighbors and reference-frame blocks are then added as safeguards. The same sparse routing rule is applied at every transformer layer.

\subsection{Camera Pose Alignment for Frustum Slab Geometry}
\label{app:future-camera-alignment}

The frustum slab table is reconstructed in the LingBot-map camera-to-world frame, while the chunk cameras used by the video generator come from ViPE poses. LingBot-map does not accept external camera poses during streaming reconstruction. \method{} therefore maps each target ViPE camera into the reconstructed coordinate frame before multiwarp, OBB construction, frustum slab IoU, and coverage estimation.

Let $(R^V_\ast,p^V_\ast)$ and $(R^L_\ast,p^L_\ast)$ denote the most recent matched ViPE and LingBot-map camera-to-world poses. The alignment fit over seen camera centers supplies a positive scale $\alpha$, and the latest pose pair defines the bridge rotation $R_b=R^L_\ast(R^V_\ast)^\top$. For a target ViPE camera $(R^V_q,p^V_q)$, the reported relative alignment constructs
\begin{equation}
    \widehat R_q = R_bR^V_q,
    \qquad
    \widehat p_q
    =
    p^L_\ast + R_b\!\left(\alpha(p^V_q-p^V_\ast)\right).
    \label{eq:relative-camera-alignment}
\end{equation}
This construction maps the latest ViPE pose exactly to the latest LingBot-map pose. It also preserves the target camera's relative rotation and scaled translation from that anchor. The frustum slab builder consumes $\widehat C_q^{-1}$, where $\widehat C_q=[\widehat R_q\mid\widehat p_q]$, as the target world-to-camera matrix. Before scoring, the implementation refreshes historical slab cameras under the current bridge so that cached geometry and target queries share one coordinate frame.

\subsection{Recency-Priority Multiwarp for Query-Block Depth}
\label{app:multiwarp-qdepth}

The current noisy chunk does not provide a reconstructed depth map, yet each query block needs a target-camera depth interval for geometric matching. Using only the latest reconstructed frame leaves holes after camera rotation or scene revisitation. Pooling all historical warps is also undesirable because stale and recent observations can place different surfaces in the same target token. Multiwarp resolves this ambiguity by selecting the newest source frame with enough support before computing any depth statistic.

\paragraph{Bounded source cache.}
For each reconstructed chunk, \method{} caches the pixel depths, camera poses, intrinsics, confidence values, and validity masks of its first and last frames. A ring buffer retains the most recent $K$ chunks, giving at most $2K$ source frames. For source frame $f$, a pixel $p$ is retained only when its depth $D_f(p)$ is finite and exceeds $\epsilon_z$, its confidence lies above the configured lower-percentile cutoff, and it passes the available static-region mask. An optional sky mask can further reject unreliable distant regions. The reported configuration uses $K=30$, $\epsilon_z=0.01$, and no sky rejection. It drops the bottom $20\%$ of confidence values among pixels with valid depth. Each cached depth map remains paired with its own LingBot-map pose, which is the camera pose under which that depth was reconstructed.

\paragraph{Warping into target latent cameras.}
Let $\mathbf{K}_f$ and $[R_f\mid\mathbf{t}_f]$ be the source intrinsics and world-to-camera pose. For a valid source pixel with homogeneous coordinate $\bar{\mathbf{p}}$, multiwarp first recovers its world point:
\begin{equation}
    \mathbf{x}^{f}_{p}=D_f(p)\mathbf{K}_f^{-1}\bar{\mathbf{p}},
    \qquad
    \mathbf{X}^{w}_{f,p}
    =
    R_f^\top\!\left(\mathbf{x}^{f}_{p}-\mathbf{t}_f\right).
    \label{eq:multiwarp-unproject}
\end{equation}
For target latent frame $t$, let $\mathbf{K}^Q_t$ and $[R^Q_t\mid\mathbf{t}^Q_t]$ denote the intrinsics and aligned world-to-camera pose from Eq.~\eqref{eq:relative-camera-alignment}. The source point becomes
\begin{equation}
    \mathbf{x}^{Q}_{f,p,t}
    =
    R^Q_t\mathbf{X}^{w}_{f,p}+\mathbf{t}^Q_t,
    \qquad
    \widetilde{\mathbf{p}}^{Q}_{f,p,t}
    =
    \pi\!\left(\mathbf{K}^Q_t\mathbf{x}^{Q}_{f,p,t}\right),
    \label{eq:multiwarp-reproject}
\end{equation}
where $\pi$ applies perspective division. Samples behind the target camera or outside the target image are discarded. If one latent token spans $h_c\times w_c$ target-image pixels, a projected pixel $(\widetilde x,\widetilde y)$ is assigned to token coordinates $(\lfloor\widetilde y/h_c\rfloor,\lfloor\widetilde x/w_c\rfloor)$. The third coordinate of $\mathbf{x}^{Q}_{f,p,t}$ supplies its target-camera depth. We write the depth multiset assigned to token $u$ as $\mathcal{Z}_{f,t,u}$ and its size as $n_{f,t,u}$.

\paragraph{Newest reliable source.}
Cached frames are ordered from oldest to newest. Multiwarp selects the newest frame that contributes at least $k_{\mathrm{warp}}$ samples to a target token:
\begin{equation}
    f^\star_{t,u}
    =
    \max\left\{f\mid n_{f,t,u}\geq k_{\mathrm{warp}}\right\}.
    \label{eq:multiwarp-winner}
\end{equation}
Only samples from $f^\star_{t,u}$ contribute to that token's statistics. The implementation records the quantile ladder
\begin{equation}
    z^\tau_Q(t,u)
    =
    \operatorname{Quantile}_{\tau}
    \!\left(\mathcal{Z}_{f^\star_{t,u},t,u}\right),
    \quad
    \tau\in\{5,10,20,50,80,90,95\},
    \label{eq:multiwarp-quantiles}
\end{equation}
and uses $z_Q(t,u)=z^{50}_Q(t,u)$ as the token depth. If the newest frame lacks enough samples, an older frame may fill the hole. If no source reaches the sample floor, the token remains invalid. Selecting the source before taking quantiles prevents depth samples from different observations and surfaces from being mixed.

\paragraph{From token depth to a query frustum slab.}
A global Gilbert query block can cross latent-frame boundaries, so \method{} first divides it into frame-local rectangular fragments. For fragment $B_{i,t}$, define its valid token set as $\mathcal{V}_{i,t}=\{u\in B_{i,t}\mid f^\star_{t,u}\ \text{exists}\}$. Its target-camera depth interval is
\begin{equation}
    z^-_{i,t}
    =
    \operatorname{Quantile}_{p_{\mathrm{lo}}}
    \!\left(\{z_Q(t,u):u\in\mathcal{V}_{i,t}\}\right),
    \qquad
    z^+_{i,t}
    =
    \operatorname{Quantile}_{p_{\mathrm{hi}}}
    \!\left(\{z_Q(t,u):u\in\mathcal{V}_{i,t}\}\right).
    \label{eq:multiwarp-block-interval}
\end{equation}
The implementation enforces $z^+_{i,t}\geq z^-_{i,t}+\epsilon_{\mathrm{slab}}$. The fragment's token rectangle and this robust interval define the query frustum slab used for world-space OBB matching. An empty $\mathcal{V}_{i,t}$ contributes no geometric evidence. If the resulting query block does not reach the coverage threshold, the router follows the model-derived path in Eq.~\eqref{eq:route-uncovered}.

\paragraph{Implementation path.}
The GPU procedure has two logical passes. The first counts valid samples for every source-frame, target-frame, and target-token tuple and selects $f^\star_{t,u}$. The second compacts only winner-frame samples and computes segmented quantiles. This ordering makes the source cache bounded, avoids cross-frame depth mixing, and supplies block-aligned Q-depth without reconstructing the noisy target chunk. The reported configuration uses $k_{\mathrm{warp}}=2$ and $(p_{\mathrm{lo}},p_{\mathrm{hi}})=(10,90)$. It does not apply the optional source-to-target pose gate or the experimental warp occluder.

\subsection{Runtime Path and Timing Scope}
\label{app:runtime-path}

The measured sparse path is a Triton indexed block-sparse attention implementation, not a Python masking wrapper around dense attention. Queries are reordered once per chunk into the global-aligned Gilbert token order, and KV caches are written directly in that layout so that the hot kernel loads contiguous KV tiles without a per-call host permutation. The kernel supports rectangular current-query versus growing-history attention, consumes the routed block lists produced by the fused-score scheduler, and uses a runtime route-length bound so later chunks do not trigger new compiler specializations. Section~\ref{app:velox-kernel} gives the full kernel path.

Frustum slab geometry is built asynchronously. After a chunk is decoded, a worker thread on a dedicated CUDA stream updates reconstruction state, records chunk cameras, builds stored frustum slabs, refreshes multiwarp frame caches, and prepares the Velox state for a later chunk while the main stream continues DiT denoising. The implementation uses a two-chunk freshness lag. Chunk $c$ normally consumes geometry through chunk $c-2$, which is sufficient because frustum slab overlap changes gradually. The worker is the sole writer to the geometry objects. The main stream waits only on a completion event when a future state is needed. Section~\ref{app:async-memory} gives the full implementation path, including the worker stream, VAE side stream, prewarming, and tail handling.

All speed numbers are measured on one H200 from the 48-chunk timing traces in Figure~\ref{fig:runtime-scaling}. We report per-chunk attention-call latency and per-chunk denoising critical-path time rather than a single aggregate wall-clock number because early chunks have small dense-attention baseline cost, while late chunks expose the scaling behavior that matters for long video generation. \rev{The denoising time is measured while the geometry worker runs on the same GPU, so it already includes the slowdown caused by sharing the device. It does not include model loading, the warm-up of the first chunks, VAE decoding, or the final drain of the worker. \method{} also keeps every historical KV block in memory, so it reduces the attention computed per chunk but not the size of the KV cache.}

\section{Asynchronous Frustum Slab Construction}
\label{app:async-memory}

The geometry pipeline would be expensive if it were inserted serially between generated chunks. In a synchronous implementation, chunk $c$ must be denoised, decoded, reconstructed into depth and pose state, inserted into the frustum slab table and multiwarp cache, aligned with the next target cameras, and scored before chunk $c+1$ can denoise. The reconstruction and frustum slab preparation path is on the same order as one DiT denoise step in our setting, so serial execution would hide the attention-kernel gains.

\method{} relaxes this dependency with a two-chunk freshness lag. Target chunk $c$ consumes a Velox state built from frustum slabs through chunk $c-2$, while the worker prepares states for later chunks. The state remains self-consistent: each worker task first inserts exactly one decoded chunk into the frustum slab table and multiwarp cache, then scores that table against the target chunk's latent cameras. Chunks before the first available Velox state run dense attention, so the method never consumes an uninitialized route.

The implementation uses one Python worker thread and one dedicated CUDA stream. The worker is the sole writer to the frustum slab table, pose aligner, multiwarp frame cache, and sparse-attention context. The main denoising thread submits a geometry task after a chunk decode and receives a future keyed by the target chunk. Immediately before chunk $c$ denoises, the main stream waits on the future's CUDA event if the worker has not already finished, then consumes the prepared geometric and importance states. This wait is a device-side stream dependency. There is no host-wide synchronization in the hot path.

Each geometry task performs four operations. First, the reconstructed frames are passed through the streaming reconstruction backend to update depths, poses, confidence, and masks. Second, the relative alignment maps target camera-control poses into the reconstructed frame and refreshes stored slab cameras under the current bridge. Third, valid depth samples are converted into per-block frustum slab primitives and inserted into the versioned frustum slab table, while first and last frame depths are cached for multiwarp. Fourth, the sparse context warps cached depths into the target cameras, builds Q frustum slabs, scores OBB overlap against historical frustum slabs, computes coverage, and materializes the geometric state consumed by soft fusion at each attention layer.

The implementation also moves VAE decode off the denoise critical path with a side stream, prewarms the worker thread's cuDNN handle, prewarms segmentation and sparse-attention kernels outside the measured interval, and skips frustum slab scoring for the final chunk because no later denoise consumes it. Frames are moved to CPU after decode and copied back on the worker stream, avoiding extra GPU residency while the main stream denoises.

All reported timing results use a single NVIDIA H200. The async worker and the DiT denoiser share that one GPU. Memory reconstruction is not offloaded to a second device. A second GPU could reduce CUDA-stream contention and further improve end-to-end time, but it is not part of the reported measurement protocol.

\section{\method{} Attention Kernel and Global Gilbert Layout}
\label{app:velox-kernel}

The sparse route is implemented as a native indexed attention path. The router produces block lists, the KV cache is stored in the order consumed by those lists, and the Triton kernel performs only the selected block interactions.

\subsection{Global-Aligned Gilbert Token Order}
\label{app:gilbert-order}

Let a latent chunk contain $T$ frames, spatial grid height $H$, spatial grid width $W$, and $S=THW$ tokens. The row-major token index is
\[
    \lambda(t,y,x) = tHW + yW + x,
    \qquad
    0 \le t<T,\ 0 \le y<H,\ 0 \le x<W.
\]
Here $t$, $y$, and $x$ are the frame, row, and column indices within the chunk. Velox uses a global-aligned Gilbert map $\Gamma:\{0,\ldots,(c+1)S-1\}\rightarrow\{0,\ldots,(c+1)S-1\}$ over the visible prefix, where $\Gamma$ maps a global row-major token index to its global Gilbert rank. The implementation builds a single-chunk Gilbert rank map $\gamma$ once, then fills each chunk with a parity-dependent reflection. If $(cT)$ is even, chunk $c$ uses the single-chunk map directly. If $(cT)$ is odd, the chunk-local coordinates are reflected by the base-frame exit quadrant before ranks are offset by $cS$. For the chunk-local reflection $F_c$, the global rank is:
\[
    \Gamma(cS+\lambda(t,y,x))
    =
    cS + \gamma\!\left(F_c(\lambda(t,y,x))\right).
\]
This incremental rule is rank-identical to building the full-sequence Gilbert curve over all visible frames. With the same token block size $b_{\mathrm{tok}}=128$ used in Section~\ref{subsec:method-prelim}, the global block id is:
\[
    \beta_{\mathrm{glob}}(c,\lambda)
    =
    \left\lfloor
    \frac{\Gamma(cS+\lambda)}{b_{\mathrm{tok}}}
    \right\rfloor .
\]

\begin{figure}[ht!]
    \centering
    \includegraphics[width=\linewidth]{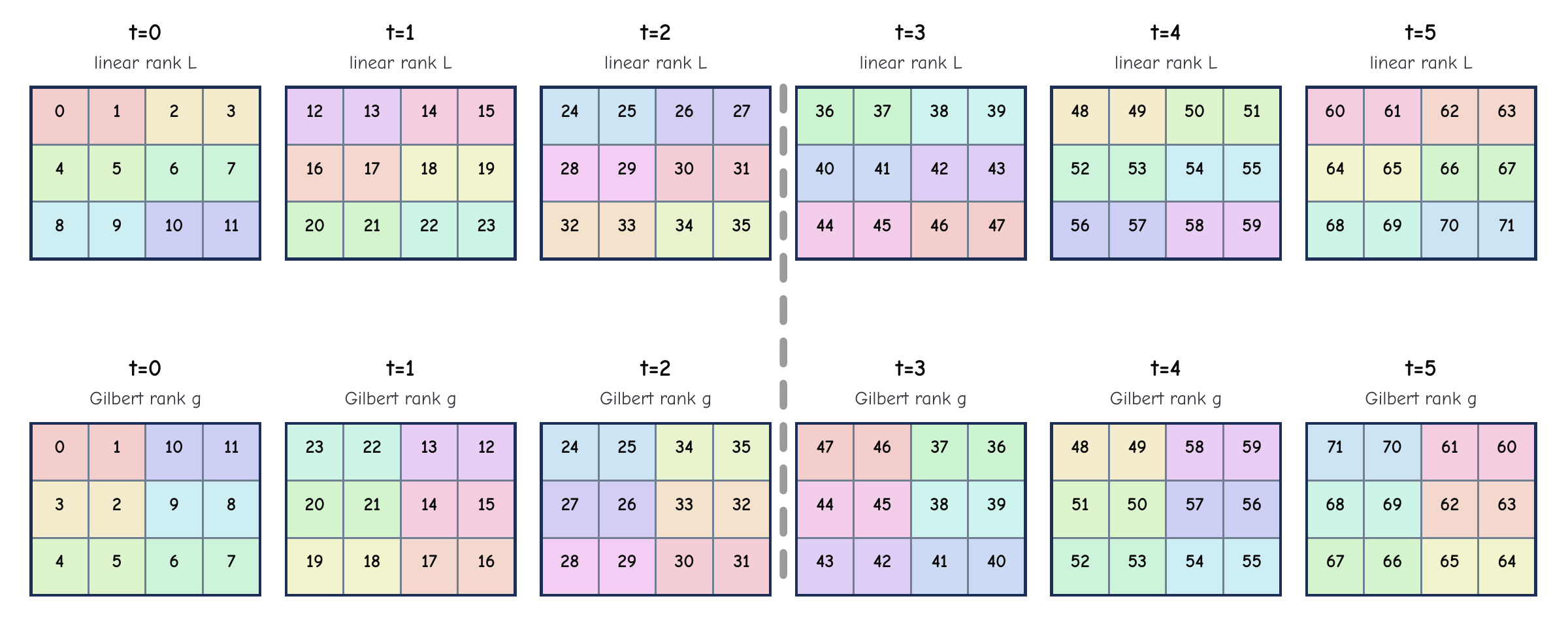}
    \caption{\textbf{THW linear order versus global Gilbert order.} Row-major THW ranking scans each frame independently, whereas a single global Gilbert curve preserves spatial locality across six frames and continues across the chunk boundary.}
    \label{fig:app-linear-vs-global-gilbert}
\end{figure}

\begin{figure}[ht!]
    \centering
    \includegraphics[width=\linewidth]{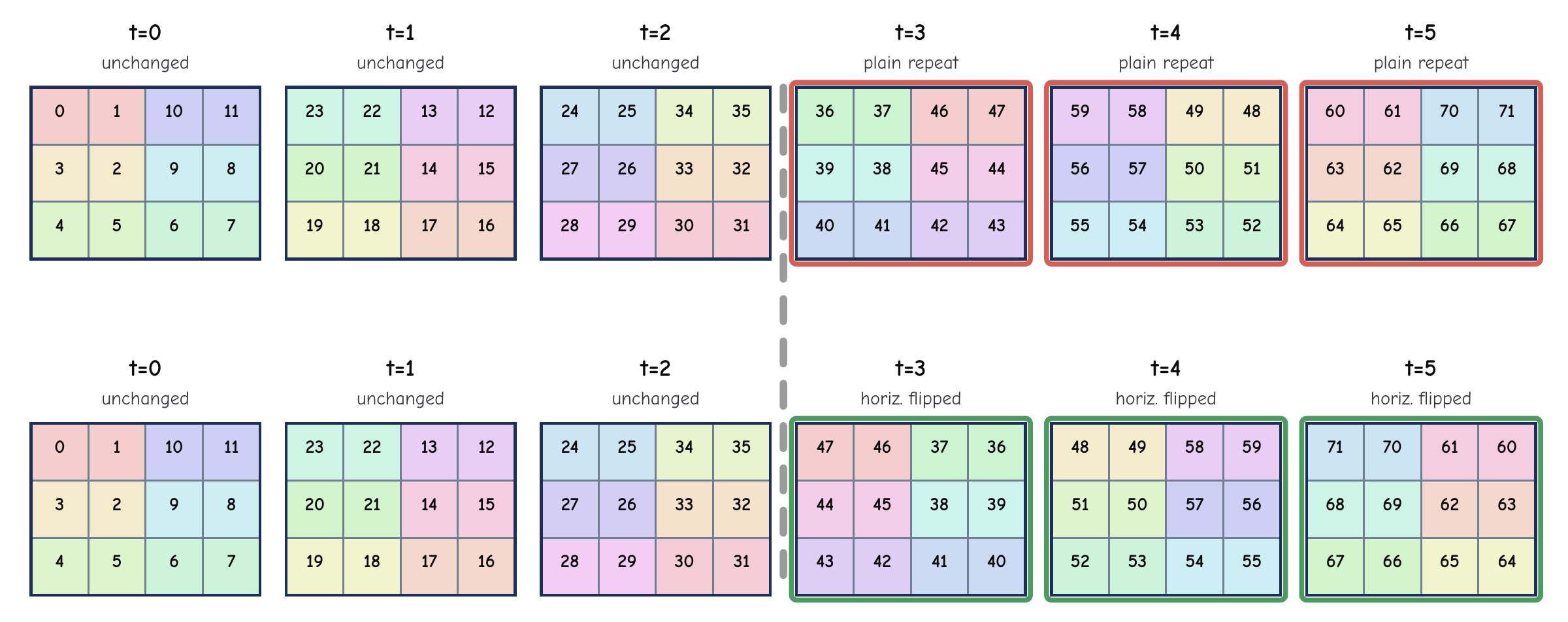}
    \caption{\textbf{Chunk-wise construction of the global Gilbert order.} Repeating the single-chunk order without reorientation creates a mismatch after the chunk boundary. Applying the parity-dependent horizontal reflection to the second chunk reproduces the corresponding segment of the full-sequence Gilbert order.}
    \label{fig:app-chunk-gilbert-flip-rule}
\end{figure}

\subsection{Global Gilbert KV Storage}
\label{app:gilbert-kv}

Velox writes each chunk's KV cache directly in its global-aligned Gilbert order. Let $\mathbf{K}_{c,\lambda,a,:}$ and $\mathbf{V}_{c,\lambda,a,:}$ denote the linear-layout key and value vectors for chunk $c$, token $\lambda$, and attention head $a$. Let $\kappa$ denote a within-chunk Gilbert rank, and let $G_c(\kappa)$ be the gather index that maps $\kappa$ back to the corresponding chunk-local linear token. The stored tensors are:
\[
    \widetilde{\mathbf{K}}_{c,\kappa,a,:}
    =
    \mathbf{K}_{c,G_c(\kappa),a,:},
    \qquad
    \widetilde{\mathbf{V}}_{c,\kappa,a,:}
    =
    \mathbf{V}_{c,G_c(\kappa),a,:},
    \qquad 0 \le \kappa < S .
\]
Each chunk slot is front-padded as needed so the current chunk's query blocks align to the global block ids in the visible prefix, and padded positions are zero-filled. The sparse kernel therefore sees a chunked contiguous address range whose block boundaries match $\beta_{\mathrm{glob}}$.

The implementation uses separate query and KV sequence lengths, because current-chunk queries attend to a growing history cache. This rectangular interface is what lets Velox replace only historical visual attention while leaving the pretrained chunked generator interface unchanged.

\subsection{Indexed Sparse Attention Kernel}
\label{app:indexed-kernel}

Fix a current chunk $c$ and layer $\ell$. Let $\mathcal{J}_i^\ell$ be the materialized block list consumed by the kernel for query block $i$. It is the implementation-level form of $\overline{\mathcal{R}}_i^\ell$ in Eq.~\eqref{eq:safeguard-route}. For the query tokens in block $i$, the kernel computes:
\[
    \mathbf{O}_{i}
    =
    \mathrm{softmax}\!\left(
    \frac{\mathbf{Q}_{i}
    \mathbf{K}_{\mathcal{J}_{i}^{\ell}}^{\top}}
    {\sqrt{d}}
    \right)
    \mathbf{V}_{\mathcal{J}_{i}^{\ell}}.
\]
The block list is different for each query block, but the route-length bound is a runtime value rather than a compiler specialization. This avoids per-chunk Triton recompilation as the retained history grows.

The kernel evaluates the softmax with online normalization. For routed KV tile step $q$ with score matrix $\mathbf{S}^{(q)}$, let $m^{(q)}$, $w^{(q)}$, and $\mathbf{A}^{(q)}$ denote the row-wise maximum, softmax normalizer, and value accumulator before processing that tile. The update is:
\[
    m^{(q+1)}
    =
    \max\!\left(m^{(q)}, \max_v \mathbf{S}^{(q)}_v\right),
    \qquad
    \alpha^{(q)}
    =
    \exp\!\left(m^{(q)}-m^{(q+1)}\right),
\]
\[
    w^{(q+1)}
    =
    \alpha^{(q)}w^{(q)}
    +
    \sum_v
    \exp\!\left(\mathbf{S}^{(q)}_v-m^{(q+1)}\right),
\]
\[
    \mathbf{A}^{(q+1)}
    =
    \alpha^{(q)}\mathbf{A}^{(q)}
    +
    \exp\!\left(\mathbf{S}^{(q)}-m^{(q+1)}\right)\mathbf{V}^{(q)} .
\]
After the final routed tile, $\mathbf{O}=\mathbf{A}/w$. In the Gilbert-storage path, KV columns are contiguous, so the kernel can use contiguous memory hints and fuse the accumulator scaling into the matrix multiply update. The gathered linear-storage path remains available for validation and ablation, but the reported timings use the contiguous Gilbert-storage path.

Velox uses a small routed-tile group in the inner loop. If $L_i=|\mathcal{J}_i^\ell|$ and $B_{\mathrm{grp}}=4$, the loop has $\lfloor L_i/B_{\mathrm{grp}}\rfloor$ full groups and $L_i\bmod B_{\mathrm{grp}}$ tail blocks. The full groups are predicate-free, and only the final tail is masked. This two-phase loop recovered the runtime-bound implementation from the scheduling overhead caused by checking the tail predicate in every unrolled group.

\subsection{Route Materialization and Caches}
\label{app:route-cache}

The route materializer reserves the previous chunk before selecting older history. For covered rows, it ranks the remaining blocks by the fused distribution:
\[
    \mathcal{J}_{i}^{\ell}
    =
    \mathrm{unique}\!\left(
    \mathcal{P}_{c-1}
    \cup
    \operatorname{Top}_{K_{\mathrm{cov}}-|\mathcal{P}_{c-1}|}
    \left(\{\widetilde p_{i,r}^{\ell}:r\notin\mathcal{P}_{c-1}\}\right)
    \cup \mathcal{L}^{\mathrm{glob}}_{i}\cup\mathcal{F}_{0}
    \right),
    \qquad \rho_i \ge \eta_{\mathrm{cov}} .
\]
For uncovered rows, it applies the same reservation to the capped model-only route:
\[
    \mathcal{J}_{i}^{\ell}
    =
    \mathrm{unique}\!\left(
    \mathcal{P}_{c-1}
    \cup
    \operatorname{Prefix}_{\eta_{\mathrm{imp}},K_{\mathrm{unc}}-|\mathcal{P}_{c-1}|}
    \left(\{p_{i,r}^{\ell}:r\notin\mathcal{P}_{c-1}\}\right)
    \cup \mathcal{L}^{\mathrm{glob}}_{i}\cup\mathcal{F}_{0}
    \right),
    \qquad \rho_i < \eta_{\mathrm{cov}} .
\]
The implementation floors the remaining selection budget to avoid an empty older-history route. The global-neighbor and reference-frame safeguards are added outside the long-history budget, which is why a materialized route can contain more than $K_{\mathrm{cov}}$ blocks.

The route builder memoizes frustum slab overlap tables, coverage, global/window local-neighbor adjacency, and pooled historical key summaries. It builds the block mask by direct indexed scatter from row and column tensors, avoiding synchronization-heavy coordinate extraction. Cross-chunk boundary neighbors are already present in the global/window 26-neighbor adjacency, so no separate seam stamping appears in the current route materialization.

\subsection{Failure Modes and Validation Scope}
\label{subsec:exp-failures}

\method{} is most reliable when the frustum slab table and multiwarp Q-depth explain the current view. The expected failure cases are therefore geometry-driven. If reconstructed depth drifts, if the multiwarp cache fills a newly disoccluded token with stale far geometry, or if OBB approximation overestimates overlap for a thin structure, soft fusion can promote stale or misleading history. The current router mitigates these cases by applying geometry only to well-covered rows, bounding the geometric multiplier, retaining the layer-specific model distribution, and preserving previous-chunk and local-neighbor safeguards. The goal is to remove history that is spatially irrelevant while retaining enough model evidence for noisy generated-video geometry.

\rev{\noindent\textbf{Camera-following errors.} The target cameras come from the commanded trajectory, while the stored slabs come from the reconstructed history. If a generated chunk does not follow its commanded camera, the warped Q-depth and the overlap scores describe a view that the model did not produce. High coverage then does not guarantee a correct match. Three parts of the design limit this error. First, the alignment in Eq.~\eqref{eq:relative-camera-alignment} anchors every target camera at the latest reconstructed pose. The error is therefore measured from the most recent generated frame rather than accumulated from the start of the rollout. Second, soft fusion only reweights the model's own importance by a factor of at most $1+\lambda=3$. A wrong overlap can therefore raise the score of a block, but it cannot force a block that the model gives little weight into the route. Third, the previous chunk and the local neighbors stay in every route, so local continuity does not depend on geometry at all. These parts reduce the damage of a wrong match but do not remove it. Appendix~\ref{app:drift} tests this failure on real rollouts and under a deliberate camera offset.}

The WBench evaluation also suggests where future quality regressions should be inspected. \rev{\method{} has higher spatial consistency and imaging quality than dense attention, but perspective and subject consistency decrease.} \rev{In the paired test of Appendix~\ref{app:paired}, the imaging gain and the perspective and subject losses are clear, while the spatial gain is within noise.} This is consistent with the design risk: static background regions are usually supported by frustum slabs, while object- or camera-sensitive regions can be dynamic, partially reconstructed, or semantically important despite weak geometric correspondence. \rev{Block-Sparse is lower than \method{} on both metrics (perspective $75.7$, subject $85.8$). This is consistent with drops caused by sparse routing itself that geometry partly recovers.}

\end{appendices}

\end{document}